\documentclass[runningheads]{llncs}

\usepackage[T1]{fontenc}
\usepackage{graphicx}
\usepackage{amsmath}
\usepackage{amssymb}
\usepackage{cite}
\usepackage{hyperref}
\usepackage{dsfont}
\usepackage{caption}
\usepackage{subcaption}
\usepackage{blindtext}
\usepackage{wrapfig}
\usepackage{colortbl}
\usepackage{xcolor}
\usepackage{booktabs}
\usepackage[letterspace=-50]{microtype}
\usepackage{multirow}

\definecolor{datasetcolor}{rgb}{0.302,0.302,0.302}
\definecolor{lightestgray}{RGB}{220,220,220}

\newcommand{\x}{\boldsymbol{x}}
\newcommand{\y}{{y}}

\newcommand{\I}{\boldsymbol{\mathrm{I}}}
\newcommand{\Bait}{\textsc{Bait}\xspace}
\newcommand{\Margin}{\textsc{Margin}\xspace}
\newcommand{\Badge}{\textsc{Badge}\xspace}
\newcommand{\Typiclust}{\textsc{Typiclust}\xspace}
\newcommand{\Random}{\textsc{Random}\xspace}

\newcommand{\param}{\theta}
\newcommand{\params}{\boldsymbol{\param}}
\newcommand{\paramsfeature}{\boldsymbol{\omega}}
\newcommand{\loglikelihood}{\ln p(\y|\x,\params)}
\DeclareMathOperator{\diag}{diag}

\newcommand{\f}{\boldsymbol{f}}
\newcommand{\h}{\boldsymbol{h}}

\begin{document}
\title{Fast Fishing: Approximating \textsc{Bait} for Efficient and Scalable Deep Active Image Classification}

\author{
Denis Huseljic\thanks{Corresponding Author} \and 
Paul Hahn \and 
Marek Herde \and 
Lukas Rauch \and 
Bernhard Sick
}
\authorrunning{
D.~Huseljic et al.
}
\titlerunning{Fast Fishing: \textsc{Bait} for Scalable Deep Active Learning}
\institute{
University of Kassel, Intelligent Embedded Systems, Germany \\
\texttt{\{dhuseljic, paul.hahn, marek.herde, lukas.rauch, bsick\}@uni-kassel.de}
}

\maketitle
\begin{abstract}
Deep active learning (AL) seeks to minimize the annotation costs for training deep neural networks. \textsc{Bait}, a recently proposed AL strategy based on the Fisher Information, has demonstrated impressive performance across various datasets. However, \textsc{Bait}'s high computational and memory requirements hinder its applicability on large-scale classification tasks, resulting in current research neglecting \textsc{Bait} in their evaluation. This paper introduces two methods to enhance \textsc{Bait}'s computational efficiency and scalability. Notably, we significantly reduce its time complexity by approximating the Fisher Information. In particular, we adapt the original formulation by i) taking the expectation over the most probable classes, and ii) constructing a binary classification task, leading to an alternative likelihood for gradient computations. Consequently, this allows the efficient use of \textsc{Bait} on large-scale datasets, including ImageNet. Our unified and comprehensive evaluation across a variety of datasets demonstrates that our approximations achieve strong performance with considerably reduced time complexity. Furthermore, we provide an extensive open-source toolbox that implements recent state-of-the-art AL strategies, available at \url{https://github.com/dhuseljic/dal-toolbox}.
\keywords{Active Learning \and Deep Learning \and Fisher Information.}
\end{abstract}

\section{Introduction}
Training deep neural networks (DNNs) requires large amounts of annotated data. In this context, the annotation process represents a significant bottleneck because it is costly and time-consuming. Active learning (AL) is an iterative process that offers a solution by intelligently selecting a subset of informative data points from an unlabeled dataset for annotation. This subset is then employed to train DNNs, reducing the need for extensive annotation processes.

In recent years, there has been a lot of progress in deep AL, with many studies focusing on developing AL selection strategies. Generally, these strategies can be categorized into uncertainty-based and diversity-based approaches. Uncertainty-based strategies leverage the uncertainty from a DNN, assuming that instances with high uncertainty are informative~\cite{gal2017deep}. However, deep AL often involves batch acquisitions since retraining DNNs is time-consuming and computationally expensive. This makes uncertainty-based strategies less effective due to the risk of picking similar instances in a batch, leading to redundancy. To address this issue, diversity-based strategies have been proposed in deep AL~\cite{sener2018active}. These prioritize acquiring diverse instances in a batch, thereby reducing information redundancy. 
Optimally, however, we aim to construct a batch that is both diverse and informative, ensuring the batch's overall informativeness is maximized. To this end, most recent studies seek to balance informativeness and diversity when selecting batches of instances~\cite{ash2020deep, ash2021gone, hacohen2022active}.

\begin{figure}[t]
    \centering
    \begin{subfigure}[b]{0.55\textwidth}
        \begin{picture}(150, 150)
            \put(10, 10){\includegraphics[height=4.9cm]{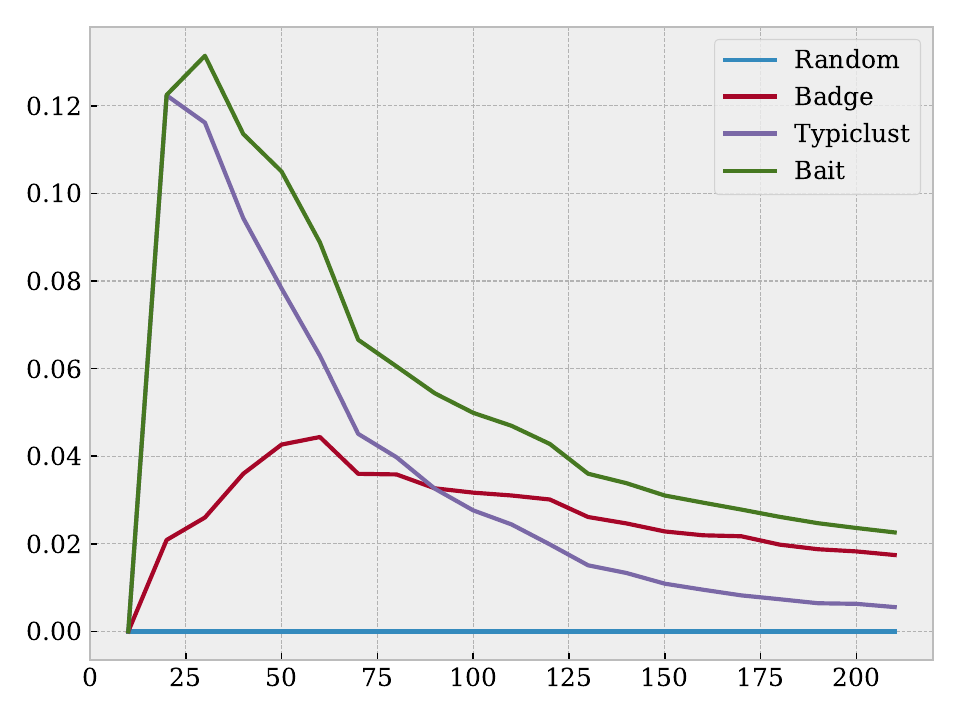}}
            \put(5, 25){\rotatebox{90}{\scriptsize Accuracy Difference to \Random}}
            \put(70, 7){{\scriptsize Number of Annotations}}
        \end{picture}
        \caption{Accuracy improvements compared to a random instance selection for popular strategies.}
        \label{fig:graphical_abstract_a}
     \end{subfigure}
     \quad
     \begin{subfigure}[b]{0.34\textwidth}
        \begin{picture}(100, 150)
            \put(10,10){\includegraphics[height=4.9cm]{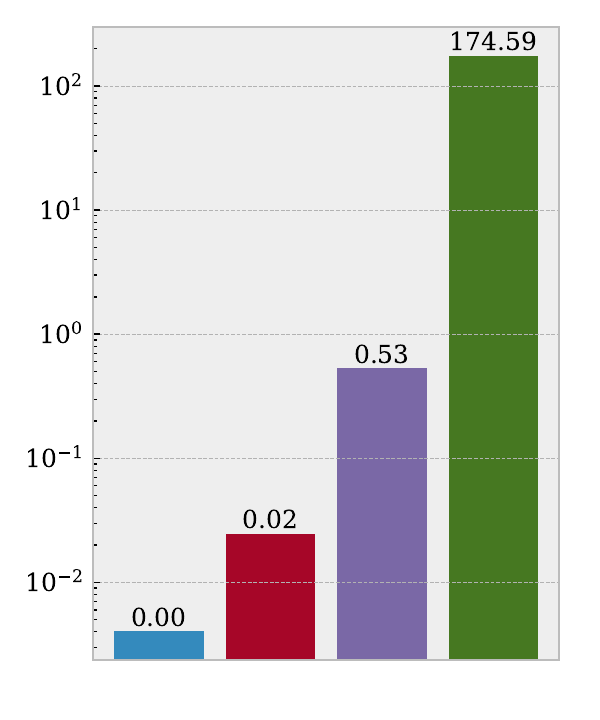}}
            \put(31, 14){{\tiny \lsstyle \Random}}
            \put(54, 14){{\tiny \lsstyle \textsc{Badge}}}
            \put(72, 14){{\tiny \lsstyle \Typiclust}}
            \put(100, 14){{\tiny \lsstyle \textsc{Bait}}}
            \put(5, 56){\rotatebox{90}{\scriptsize Acquisition Time}}
        \end{picture}
        \caption{Average acquisition time per cycle in seconds.}
        \label{fig:graphical_abstract_b}
     \end{subfigure}
    \caption{Comparison of different AL strategies on CIFAR-10.}
    \label{fig:graphical_abstract}
\end{figure}

In the plethora of those deep AL strategies, \textsc{Bait}~\cite{ash2021gone} stands out as an approach that offers superior performance. 
Central to \textsc{Bait} is optimizing the Bayes risk through an objective function that requires the calculation of the Fisher Information Matrix (FIM). Since the calculation of the FIM in high-dimensional parameter spaces is infeasible, \textsc{Bait} uses the parameters of the DNN's last layer. Empirically, it shows impressive results across different datasets~\cite{chen2023firal}. Despite its merits, many recent studies neglect to compare their strategy against \textsc{Bait}~\cite{hacohen2022active,yehuda2022active,tan2023bayesian,rauch2023activeglae}. One reason may be the poor time complexity of $\mathcal{O}(D^2K^3)$ when computing the FIM for a large number of classes $K$ and dimensions $D$, or the high time and space requirements for the selection based on an FIM per instance. Additionally, the lack of accessible implementations and the complexity of integrating \textsc{Bait} into existing frameworks could prevent its broad acceptance.

Figure~\ref{fig:graphical_abstract} presents the results of an AL experiment on CIFAR-10 with a total budget of 210 instances and a batch acquisition size of 10 instances per cycle. The detailed experimental setup is available in the Appendix~\ref{app:graphical_abstract}. We evaluate prominent state-of-the-art AL strategies, focusing on performance (i.e., accuracy) and acquisition times (i.e., required time for selecting instances per cycle). Fig.~\ref{fig:graphical_abstract_a} plots the accuracy difference between a model trained with an AL strategy and a model trained with randomly sampled instances. Notably, \textsc{Bait} outperforms every strategy, demonstrating its superior performance. However, the acquisition times in Fig.~\ref{fig:graphical_abstract_b} indicate that \textsc{Bait} requires significantly more time than others to acquire a batch of instances. This bottleneck will become even more severe when dealing with larger acquisition sizes or more classes. Consequently, despite \textsc{Bait}'s superior accuracy, this example highlights its computational challenges. These practical constraints limit its feasibility across various applications and real-world settings.

For this reason, we aim to improve \textsc{Bait}'s computational efficiency and scalability across classes by introducing two methods to tackle its main bottlenecks.
Our first method focuses on an efficient computation of the FIM, improving the time complexity to $\mathcal{O}(D^2 K^2)$ by only focusing on the most probable classes in the expectation.
Our second method focuses on reducing the size of the FIM, leading to a time complexity of $\mathcal{O}(D^2)$. 
\textsc{Bait}'s original formulation requires the calculation of an FIM per instance, posing challenges in computational and memory resources, especially as the number of entries in the FIM increases quadratically with more classes.  
To overcome this, we propose to approximate the original multi-class likelihood through a binary one that significantly lowers the dimensionality, leading not only to an efficient computation of the FIM but also an efficient computation of \textsc{Bait}'s objective. This way, we enable scalability across many classes while enhancing computational efficiency. While the first method prioritizes adherence to the original FIM formulation, the second approach aims to completely disentangle time and space complexity from the number of classes, enabling \textsc{Bait}'s usage on large-scale datasets such as ImageNet~\cite{deng2009imagenet}.
Our contributions can be summarized as follows:
\begin{itemize}
    \item We propose two approximation methods for \textsc{Bait}, enhancing its computational efficiency and scalability to many classes without compromising performance.
    \item We provide a unified and comprehensive study comparing recent state-of-the-art AL strategies across a multitude of image datasets.
    \item We provide a toolbox that implements recent state-of-the-art strategies, including our version of \textsc{Bait}, for a straightforward adoption in future research.
\end{itemize}

\section{Related Work}\label{sec:related_work}
Pool-based \textbf{deep AL strategies} select batches instead of single instances to avoid retraining an DNN with each new acquired instance~\cite{ren2020survey}. AL strategies are typically categorized into uncertainty-based, diversity-based, and hybrid strategies. Uncertainty-based strategies aim to select instances that are assumed to be difficult for the DNN, often close to the decision boundary. Employed in a batch acquisition setting, uncertainty-based strategies greedily select the highest-scoring instances, which can lead to information redundancy. \Margin sampling~\cite{settles2009active}, a variant of uncertainty sampling, selects instances where the difference between the two highest predicted class probabilities is largest. Recently, it has been shown to work effective in AL~\cite{bahri2022margin}, despite information redundancy in a batch. \textsc{BALD}~\cite{gal2017deep} assumes a Bayesian DNN and selects instance that maximize the mutual information between predictions and the DNN's posterior distribution. BatchBALD~\cite{kirsch2019batchbald} improves the selection of BALD by reducing information redundancy in a batch.
In contrast to uncertainty-based strategies, diversity-based strategies aim to select a batch of diverse instances that represent the dataset.
CoreSet~\cite{sener2018active} selects instances that minimize the average distance between labeled and unlabeled instances.
Recently, there has been a rise in hybrid strategies that combine the uncertainty and diversity-based selection.
\Badge~\cite{ash2020deep} selects via k-means++~\cite{arthur2007k} on instances represented in a gradient space, which results in a selection of difficult and diverse instances.
Bayesian Estimate of Mean Proper Scores~\cite{tan2023bayesian} estimates the loss reduction in combination with k-means to ensure diversity.
\Typiclust~\cite{hacohen2022active} focuses on instances with high density through nearest neighbor density estimation. They employ k-means to ensure diversity.

The Fisher information describes the amount of information an observable random variable carries about an unknown parameter. It is commonly used as an approximation to the Hessian in the field of optimization~\cite{amari1998natural}. In the context of AL, \cite{chaudhuri2015convergence} motivate its use by the selection of instances that minimize the expected likelihood given the current set of parameters. They propose a near-optimal strategy consisting of two phases, which require multiple calculations of the FIM. Since the size of the FIM scales quadratically with the number of parameters, employing this scheme with DNNs is infeasible. Therefore, \Bait~\cite{ash2021gone} proposes several adaptions, discussed in Section~\ref{sec:bait}, to combine this strategy with DNNs. 
In the context of second-order optimization, there exist several works focusing on the approximation of the FIM. \cite{lecun1989optimal} propose to reduce the FIM to its diagonal elements, effectively decreasing the quadratic scaling in the number of parameters to a linear scaling. \cite{heskes2000natural} assume independency between parameter groups and, therefore, reduce the FIM to a block diagonal matrix. \cite{martens2015optimizing} propose the Kronecker Factorized Approximate Curvature, which approximates blocks of the FIM as the Kronecker product of two smaller matrices, leading to computational savings in both computation and inversion. While \cite{sourati2019intelligent} reformulate the score function of the FIM as averages over parameters, \cite{lee2020estimating} improve the efficiency of KFAC by using dimensionality reduction techniques such as low-rank approximations.
For a comprehensive comparison regarding these approximations, we refer to \cite{daxberger2021Laplace}. For a more detailed theoretical discussion about the FIM, we refer to \cite{kunstner2019limitations}.

\section{Notation}
We focus on a pool-based classification setting with instances $\x \in \mathcal{X}$ and associated class labels $\y \in \mathcal{Y} = \{1, \dots, K\}$, where $K$ is the number of classes. 
A DNN consists of a feature extractor and a classification layer. The former is given by $\h_{\paramsfeature}: \mathcal{X} \to \mathbb{R}^D$, with parameters $\paramsfeature$, mapping the inputs to a hidden representation. The latter is given by $\f_{\params}: \mathbb{R}^D \to \mathbb{R}^K$, with parameters $\params$, mapping the hidden representation to a logit space. Class probabilities are obtained via the softmax function $p_{\params}(\y | \x) = [\operatorname{softmax}(f_{\params}(\h_{\paramsfeature}(\x)))]_\y$.
An AL cycle starts with a labeled pool $\mathcal{L} = \{(\x_n, \y_n)\}_{n = 1}^{N}$ and a large unlabeled pool $\mathcal{U} = \{\x_m\}_{m=1}^{M}$, where $N$ and $M$ denote the number of labeled and unlabeled instances, respectively. 
With this setting, we train a DNN on $\mathcal{L}$, assess the informativeness of instances in $\mathcal{U}$, query an oracle for labels on a batch of $B \in \mathbb{N}$ informative instances, add these annotated instances to $\mathcal{L}$ and remove them from $\mathcal{U}$, and repeat the cycle.  Hence, across AL cycles, the cardinality $N$ and $M$ of the sets $\mathcal{L}$ and $\mathcal{U}$ are consistently changing. Assuming instance are independent and identically distributed (i.i.d.), we typically train a DNN maximizing ${L}(\params) = \sum_{n = 1}^N \ln p(\y | \x, \params)$, where $\ln p(\y | \x, \params)$ is the likelihood function.

\section{Time and Space Complexity of \textsc{Bait}}\label{sec:bait}
The general idea of \textsc{Bait} is to optimize the Bayes risk by adapting the near-optimal two-phase sampling scheme proposed in~\cite{chaudhuri2015convergence} for the use with DNNs. For this purpose, they introduce three key extensions: (1) focusing on the DNN's last layer due to the high-dimensional parameter space, (2) recalculating the FIM after each retraining to account for changing representations, and (3) developing a greedy method for batch selection. In principle, \textsc{Bait} selects instances based on an informativeness score that is defined by 
\begin{align}
    \arg\min_{\x \in \mathcal{U}} \mathrm{tr}\left((\mathbf{M} + \I(\x; \params))^{-1} \I(\params)\right), \label{eq:info_score}
\end{align}
where $\mathbf{M} \in \mathbb{R}^{DK \times DK}$ is the FIM of already selected instances (i.e., labeled pool and instances in an acquisition batch), $\I(\params)\in \mathbb{R}^{DK \times DK}$ is the FIM of all instances (i.e., labeled and unlabeled pool), and $\I(\x; \params)\in \mathbb{R}^{DK \times DK}$ is the FIM of a particular instance $\x \in \mathcal{U}$. 
Note that the computation of the FIM does not require labels. For a detailed derivation and description, we refer to the original work~\cite{ash2021gone}. 

We see that the informativeness score in Eq.~\eqref{eq:info_score} consists of the three FIMs, and hence, the main computations revolve around \textit{calculating} these three matrices. Generally, the FIM reflects the amount of information that the random variable $\y$ carries about the parameters $\params$ given a likelihood function $\loglikelihood$. It is defined as the expected outer product of the likelihood's gradient and given by
\begin{align}
    \I(\x;\params) &= \mathbb{E}_\y[  \nabla_{\params} \ln p(\y | \x, \params) \nabla_{\params}^\mathrm{T} \ln p(\y | \x, \params)],\label{eq:fim}
\end{align}
where the expectation is taken over the categorical distribution $y \sim \text{Cat}(\y| p_{\params}(\cdot | \x))$.
The FIMs, $\I(\params)$ and $\mathbf{M}$, are obtained by averaging $\I(\x;\params)$ over the relevant instances.

We aim to reduce \textsc{Bait}'s computational time and memory footprint by approximating Eq.~\eqref{eq:fim} efficiently.  Consequently, this simplifies the calculation of the informativeness score from Eq.~\eqref{eq:info_score}, as all three matrices are calculated more efficiently. Since the outer product has a time complexity of $(K D)^2$ and we need to calculate it repeatedly for each class due to the expectation, we end up with a time complexity of $\mathcal{O}(K (K D)^2)$. As we see, the complexity of this task stem mainly from the \textbf{expectation} over the categorical distribution and the dimensionality of the likelihood's \textbf{gradient}. 

Considering this expectation, we realize that it leads to a cubic time complexity regarding $K$, which hinders \textsc{Bait} from being used in a setting with many classes. Moreover, we have these computational costs for every instance within the unlabeled and labeled pool.
In addition, the computation of the expectation not only affects time complexity but also drastically increases the memory requirements, which can lead to problems, especially when working with GPUs. 
Originally, \textsc{Bait} avoids storing a separate FIM for each instance to mitigate memory constraints. Instead, it retains only the gradients for each instance and class, thus reducing the space complexity from $\mathcal{O}(M (K D)^2)$ to $\mathcal{O}(M K (K D) )$. However, this method does not fully solve memory inefficiencies.

The dimensionality of gradients further complicates these problems. Increasing the number of classes not only increases the amount of gradients to be stored (first $K$) but also increases the dimensionality of those gradients ($K D$).
This is due to the parameters in the last layer increasing linearly with $K$ as the number of classes grows. As a result, the number of elements that need to be stored grows quadratically with the number of classes (expectation and gradient's dimensionality), which is a major challenge for applying \textsc{Bait} to large-scale problems. In addition, the outer product in Eq.~\eqref{eq:fim} leads to a high-dimensional FIM, which aggravates subsequent operations such as inversions and thus also affects the overall efficiency.

\section{Approximations}
In the previous section, we analyzed the time and space complexity of \textsc{Bait}. We demonstrated that the expectation and dimensionality of the likelihood's gradient pose a significant challenge to \textsc{Bait}'s practical applicability. Here, we present our modifications of \textsc{Bait} to solve these problems by reducing time and space complexity based on approximations of the FIM.
In our first approach, we prioritize adherence to the original objective. Conversely, our second strategy aims to reduce the dependency on $K$, resulting in a distinctly different reformulation of the FIM itself.
We refer to the first approximation as \textsc{Bait} (Exp) and to the second as \textsc{Bait} (Binary).

\subsection{Expectation}\label{sec:approx_expectation}
Our first method reduces the complexity by focusing on the expectation. This expectation is taken over the distribution $\text{Cat}(\y| p_{\params}(\cdot| \x))$ and, intuitively, it weighs the outer products by the model's predicted probability for a respective class:
\begin{align}
    \I(\x;\params) = \sum_{\y \in \mathcal{Y}} p_{\params}(\y | \x) \left( \nabla_{\params} \ln p(\y | \x, \params) \nabla_{\params}^\mathrm{T} \ln p(\y | \x, \params) \right).
\end{align}
Our idea is that, instead of taking the expectation over the complete categorical distribution, we approximate it by considering only a subset of classes. The motivation is that for many classes, most of the probability mass will be present in the top predictions of our model, leading to a good approximation of the true FIM. This way, we only have to specify the number of top classes that should be considered in the expectation.
Formally, this can be seen as taking the expectation over a new categorical distribution, which now only considers the top predictions instead of all classes. To ensure that this categorical distribution and, hence, the FIM is well defined, we additionally need to normalize the selected probabilities. 
As a result, the expectation is calculated on
\begin{align}
    \text{Cat}(\hat{y}|\hat{p}_{\params}(\cdot | \x, c)) \quad \text{with} \quad \hat{p}_{\params}(\hat{\y} | \x, c) =  \frac{p_{\params}(\hat{\y} | \x)}{\sum_{\y^\prime \in \mathcal{Y}_c} p_{\params}(\y^\prime | \x)},
\end{align}
where $\hat{y} \in \mathcal{Y}_c$ and $\mathcal{Y}_c \subset \mathcal{Y}$ denotes the $c \in \mathbb{N}_{>0}$ top-predictions of the DNN with parameters $\params$ for instance $\x$. 
This approximation effectively reduces the time complexity from $\mathcal{O}(K (K D)^2))$ to $\mathcal{O}(c (K D)^2))$, where $c$ is a constant factor unaffected by the overall number of classes. Moreover, it lowers the space complexity from $\mathcal{O}(M  D  K^2)$ to $\mathcal{O}(M  D  c  K)$, effectively diminishing the quadratic growth of memory requirements.

When considering all classes ($c = K$), we retrieve the original FIM formulation as specified in Eq.~\eqref{eq:fim}. In contrast, for a single class ($c=1$), at first glance, our method yields an approximation seemingly akin to the empirical FIM. The empirical FIM is similar to the true FIM but avoids the calculation of expectations by directly leveraging available labels $\y$. 
Since it offers better time and space complexity by avoiding the expectation, it is frequently employed as a replacement for the true FIM~\cite{kingma2014adam, george2018fast, zhang2018noisy}. However, many studies criticize its use~\cite{kunstner2019limitations, pascanu2013revisiting, martens2020new}. From a theoretical perspective, the empirical FIM is not an approximation and does not possess the properties of the true FIM. 
Comparing the empirical FIM to our approximation, we employ the DNN's prediction instead of the label.
As mentioned by~\cite{kunstner2019limitations}, the formulation we employ is not equivalent to the empirical FIM, but a biased estimate of the true FIM.
An unbiased estimate can be obtained through sampling from $\text{Cat}(\y| p_{\params}(\cdot| \x))$, albeit at the expense of increased variance~\cite{kunstner2019limitations}.
Nonetheless, within the context of AL, we empirically found that adopting a biased estimate of the FIM in favor of reduced variance proved advantageous. 
For a more detailed discussion of this topic, we refer to~\cite{kunstner2019limitations}.

\begin{table}[t]
    \centering
    \setlength{\tabcolsep}{4pt}
    \caption{Summary of time and space complexity of the proposed approximations.}
    \begin{tabular}{|l|c|c|}
    \toprule
      & \textbf{Time Complexity} & \textbf{Space Complexity} \\ 
    \midrule
    \textsc{Bait} & $\mathcal{O}(K (K D)^2)$  & $\mathcal{O}(M  D  K^2)$   \\
    \textsc{Bait} (Exp)          & $\mathcal{O}(c (K D)^2)$  &  $\mathcal{O}(M  D  c  K)$  \\
    \textsc{Bait} (Binary)          & $\mathcal{O}(D^2)$    & $\mathcal{O}(M D)$        \\
    \bottomrule
    \end{tabular}
    \label{tab:approx-complexities}
\end{table}

\subsection{Gradient}\label{sec:approx_gradient}
Our second method aims to decouple the time and space complexity of computing the FIM from the number of classes. This requires a class-independent formulation for both the expectation and gradient calculations. Consequently, we explore an alternative representation of the FIM, conceptualized via the Hessian of the likelihood function as follows:
\begin{align}
    \I(\x; \params) = - \mathbb{E}_\y[  \nabla^2_{\params} \ln p(\y | \x, \params)].
\end{align}
This formulation interprets the FIM as the negative expectation of the Hessian with respect to the model parameters. Similar to the original objective function from Eq.~\eqref{eq:fim}, we notice that many classes lead to quadratic growth of the Hessian's dimensions.
Our key idea involves approximating the Hessian and, consequently, the FIM by adopting a different likelihood $\loglikelihood$. This approach is inspired by~\cite{liu2023simple}, where class-specific covariance matrices of a Laplace approximation are approximated by considering an upper bound. This upper bound transforms the multi-class classification setting into a binary one, where the maximum probability is considered the positive class' probability. This way, we assume a shared Hessian matrix across classes as a simplification, which significantly reduces the time complexity to $\mathcal{O}(D^2)$ and space complexity to $\mathcal{O}(M D)$.
Formally, we replace the categorical likelihood with a Bernoulli likelihood, which is given by
\begin{align}
    \loglikelihood = y \ln \hat{p}  + (1 - y) \ln (1 - \hat{p})
\end{align}
where $\hat{p} = \max_\y(p_{\params}(\y | \x))$ is the highest predicted probability of our DNN. Intuitively, as we basically assume a binary classification setting, the FIM only consider the influence of a subset of parameters that led to the highest predicted probability.
When comparing the first approximation \textsc{Bait} (Exp) with \textsc{Bait} (Binary), the main difference lies in the modification of the likelihood. This adaption not only ensures better time complexity of the expectation, but also reduces the dimensionality of the gradient $\nabla_{\params} \loglikelihood$ considerably, making it independent of the number of classes.
A summary of time and space complexity of the proposed approximations is given in Tab.~\ref{tab:approx-complexities}.

\section{Experimental Results}\label{sec:experiments}
In this section, we investigate the performance of the proposed approximations by comparing them to the original version of \textsc{Bait} and applying them to real-world tasks across various image datasets. 

\subsection{Setup}
\textbf{Datasets:} 
We conduct experiments across a variety of image datasets to assess the effectiveness and robustness of our proposed approximations of \textsc{Bait}. Our evaluation encompasses a total of nine image datasets. Aiming to ensure a comprehensive evaluation focusing on scalability, we cover a broad range of class numbers. These datasets range from a small (CIFAR-10~\cite{krizhevsky2009learning}, STL-10~\cite{wang2016unsupervised}, and Snacks~\cite{snacks2023dataset}) to a large amount of classes (CIFAR-100~\cite{krizhevsky2009learning}, Food-101~\cite{bossard2014food}, Flowers-102~\cite{nilsback2008automated}, StanfordDogs~\cite{zhao2020universal}, TinyImageNet~\cite{le2015tiny}, ImageNet~\cite{deng2009imagenet}). For each dataset, we use a 10\% split of the training dataset for development of our approximations. 
Information about the number of instances and classes is summarized in Tab.~\ref{tab:dataset}. A more comprehensive description can be found in Appendix~\ref{app:datasets}. 

\begin{wraptable}{r}{5.5cm}
\vspace{-2em}
\setlength{\tabcolsep}{1pt}
\centering
\scriptsize
\caption{Image datasets overview.}
\begin{tabular}{|c|c|c|c|}
\toprule
\textbf{Dataset}  & |\textbf{Train}| & |\textbf{Test}| & |\textbf{Classes}| \\ 
\midrule
CIFAR-10        & 50k   & 10k   & 10    \\
STL-10          & 5k    & 8k    & 10    \\
Snacks          & 5k    & 1k    & 20    \\
CIFAR-100       & 50k   & 10k   & 100   \\
Food-101        & 75k   & 25k   & 101   \\
Flowers102      & 1k    & 6k    & 102   \\
StanfordDogs    & 18k   & 2k    & 120   \\
Tiny ImageNet   & 100k  & 10k   & 200   \\
ImageNet        & 1.2m  & 100k  & 1000  \\
\bottomrule
\end{tabular}
\vspace{-2em}
\label{tab:dataset}
\end{wraptable}

\textbf{Model:} We employ a Vision Transformer (ViT)~\cite{dosovitskiy2020image} with pretrained weights obtained through self-supervised learning together with a randomly initialized fully connected layer. Specifically, we use the DINOv2-ViT-S/14 model~\cite{oquab2023dinov2} with approximately 14 million parameters and a feature dimension of $D = 384$ in its final hidden layer. We intentionally select the worst-performing DINO model to simulate real-world deep AL scenarios where optimal pre-trained models may not be readily available. This setting aligns with recent recommendation from the literature~\cite{hacohen2022active} that highlight the importance of proper feature representation in deep AL for images. In each AL cycle, we train the DNN's randomly initialized last layer for 200 epochs, employing the Rectified Adam optimizer~\cite{liu2019variance}. We use a training batch size of 128, a learning rate of 0.2, and weight decay of 0.0001. In addition, we utilize a cosine annealing learning rate scheduler. We determined these hyperparameters empirically to be effective across all datasets by evaluating the convergence of a DNN on randomly sampled instances.

\textbf{AL Setting:}
We determine the acquisition size, total budget, and the number of initial instances according to the dataset characteristics to guarantee convergence of the AL process. Additionally, the initial labeled pool is chosen randomly. Details are provided in Appendix~\ref{app:al_settings}.
Our evaluation comprises various AL selection strategies, including \Random, which selects instances randomly, \Margin~\cite{bahri2022margin}, focusing on the uncertainty in the top-2 predicted probabilities, \Badge~\cite{ash2020deep}, which selects instances according to the gradient norm of the likelihood, and \Typiclust~\cite{hacohen2022active}, employing clustering and density estimation for selection. These strategies are considered state-of-the-art and were chosen based on their robustness and efficiency across different domains~\cite{bahri2022margin,rauch2023activeglae,hacohen2022active}.
Additionally, we employ \textsc{Bait} with our expectation approximation, called \textsc{Bait} (Exp), and our binary approximation, called \textsc{Bait} (Binary).

\textbf{Metrics:}
To evaluate the effectiveness of a strategy, we consider its learning curve. Therefore, in each cycle, we train a DNN on a labeled pool selected by a given strategy and compute the accuracy on the test dataset. Then, we examine the accuracy improvement over \Random by considering the difference of learning curves. Due to space limitations, we only report a selection of learning curves. All remaining curves of all strategies and datasets can be found in Appendix~\ref{app:learning_curves}. As an alternative, we report the area under the learning curve (AUC), which summarizes the learning curve into a single numerical value. All learning curves and numerical values are averaged over ten repetitions to ensure comparability.

\subsection{Assessment of Approximations}
We compare our approximations to the original \textsc{Bait} strategy across different image datasets. Due to the memory limitations of the original \textsc{Bait} formulation (cf.~Tab.~\ref{tab:approx-complexities}), it was infeasible to include datasets with more than 20 classes in these experiments. Therefore, we use CIFAR-10, STL-10 and Snacks to assess whether our approximations perform similarly to the original version.
We report acquisition times per AL cycle for a batch size of 10 instances, measured on both CPU and GPU. These times were determined using a workstation equipped with an NVIDIA RTX 4090 GPU and an AMD Ryzen 9 7950X CPU.

\begin{table}[t]
    \setlength{\tabcolsep}{4pt}
    \centering
    \caption{Comparison of \textsc{Bait} (Exp) to the original formulation of \textsc{Bait} with varying values for $c$ over multiple datasets.}
    \scriptsize
    \begin{tabular}{|l|c|c|c|c|c|}
    \toprule
    & \multirow{2}{*}{\Random}
    & \multicolumn{1}{c|}{\textsc{Bait} (Exp)}  
    & \multicolumn{1}{c|}{\textsc{Bait} (Exp)} 
    & \multicolumn{1}{c|}{\textsc{Bait} (Exp)} 
    & \multicolumn{1}{c|}{\multirow{2}{*}{\textsc{Bait}}} \\
    
    & 
    & $c=1$
    & $c=2$
    & $c=5$
    &  \\    
    \midrule
    \hline
    \multicolumn{6}{|c|}{\cellcolor{datasetcolor!10} CIFAR-10 (10 Classes)} \\
    \hline
    Accuracy (AUC)             & 82.05 & +4.63 & +5.39 & \bf{+5.50} & +5.37 \\
    \hline
    Acquisition Time (CPU)     & 00:00 & 18:07 & 18:16 & 22:44 & 32:07 \\
    Acquisition Time (GPU)     & 00:00 & 00:41 & 00:45 & 01:00 & 01:23 \\
    \hline
    \multicolumn{6}{|c|}{\cellcolor{datasetcolor!10} STL10 (10 Classes) } \\
    \hline
    Accuracy (AUC)             & 85.79 & +5.61 & +6.19 & +6.45 & \bf{+6.54} \\
    \hline
    Acquisition Time (CPU)     &  00:00 & 01:47 & 01:49 & 02:15 & 03:10 \\
    Acquisition Time (GPU)     &  00:00 & 00:04 & 00:04 & 00:06 & 00:08 \\
    \hline
    \multicolumn{6}{|c|}{\cellcolor{datasetcolor!10} Snacks (20 Classes)} \\
    \hline
    Accuracy (AUC)             & 71.93 & +7.50 & \bf{+8.44} & +8.39 & +8.07 \\
    \hline
    Acquisition Time (CPU)     &  00:00 & 09:30 & 10:40 & 12:40 & 21:50  \\
    Acquisition Time (GPU)     &  00:00 & 00:15 & 00:15 & 00:20 & 00:40  \\ 
    \bottomrule
    \end{tabular}
    \label{tab:ablation_expectation}
\end{table}
\textbf{\textsc{Bait} (Exp):} First, we examine the initial approximation discussed in Section \ref{sec:approx_expectation}, which estimates the expectation by focusing on the most probable classes. Intuitively, increasing the number of considered classes should yield an estimate closer to the true FIM. According to Table \ref{tab:ablation_expectation}, most of our approximations ($c > 1$) closely match the performance of the original \textsc{Bait} formulation across various datasets while reducing acquisition time. Most importantly, this time reduction becomes more substantial when dealing with a larger number of classes, as seen by the acquisition times on the Snacks dataset. In this case, our approximation with $c=2$ is able to cut \textsc{Bait}'s acquisition time down to a half while providing a better accuracy than the original formulation. Furthermore, in two out of three data sets, our approximations manage to outperform the original \textsc{Bait}, demonstrating that focusing on the most probable class can even be beneficial for the selection of \textsc{Bait}.

When considering a single class only in our approximation ($c = 1$), the performance decreases slightly, indicating that a single class may not be enough for an accurate approximation of the FIM. Despite that, this approximation still outperforms \Random in terms of accuracy while improving \textsc{Bait}'s time complexity. Furthermore, we already achieve comparable performances to the original \textsc{Bait} formulation with $c = 2$, suggesting that this number may be sufficient for an accurate approximation. Thus, in the remainder of the experiments, we will fix $c=2$ for \textsc{Bait} (Exp).

\begin{table}[t]
    \setlength{\tabcolsep}{4pt}
    \centering
    \scriptsize
    
    \caption{Comparison of \textsc{Bait} (Binary) and \textsc{Bait} (Diagonal) to the original formulation of \textsc{Bait} over multiple datasets.}
    \begin{tabular}{|l|c|c|c|c|}
    \toprule
    & \multirow{2}{*}{\Random}
    & \multicolumn{1}{c|}{\multirow{2}{*}{\textsc{Bait} (Diagonal)}} 
    & \multicolumn{1}{c|}{\multirow{2}{*}{\textsc{Bait} (Binary)}}
    & \multicolumn{1}{c|}{\multirow{2}{*}{\textsc{Bait}}} \\
    & 
    & 
    &
    &  \\  
    \midrule
    \hline
    \multicolumn{5}{|c|}{\cellcolor{datasetcolor!10} CIFAR-10 (10 Classes)} \\
    \hline
    Accuracy (AUC)               & 82.05 & +1.25 & \bf{+5.74} & +5.23 \\
    \hline
    Acquisition Time (CPU)  & 00:00 & 01:23 & 00:13 & 32:07 \\
    Acquisition Time (GPU)  & 00:00 & 00:32 & 00:09 & 01:23 \\
    \hline
    \multicolumn{5}{|c|}{\cellcolor{datasetcolor!10} STL10 (10 Classes) } \\
    \hline
    Accuracy (AUC)               & 85.79 &  +3.43 & +6.45 & \bf{+6.54} \\
    \hline
    Acquisition Time (CPU)  & 0.00  &  00:08  & 00:01  & 03:10  \\
    Acquisition Time (GPU)  & 0.00  &  00:03  & 00:01  & 00:08 \\
    \hline
    \multicolumn{5}{|c|}{\cellcolor{datasetcolor!10} Snacks (20 Classes)} \\
    \hline
    Accuracy (AUC)               & 71.93  & +0.77 & \bf{+8.60}  & +8.07 \\
    \hline
    Acquisition Time (CPU)  &  00:00  & 00:32 & 00:01  &  21:50 \\
    Acquisition Time (GPU)  &  00:00  & 00:11 & 00:01  &  00:41 \\
    \bottomrule
    \end{tabular}
    \label{tab:ablation_likelihood}
\end{table}

\begin{figure}[t]
    \centering
    \begin{picture}(300, 20)
    \put(25, 0){\includegraphics[width=.8\linewidth]{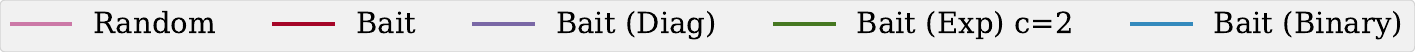}} \end{picture}
    \\
    \begin{subfigure}[b]{0.31\textwidth}
        \begin{picture}(100, 120)
            \put(10, 10){\includegraphics[width=\linewidth]{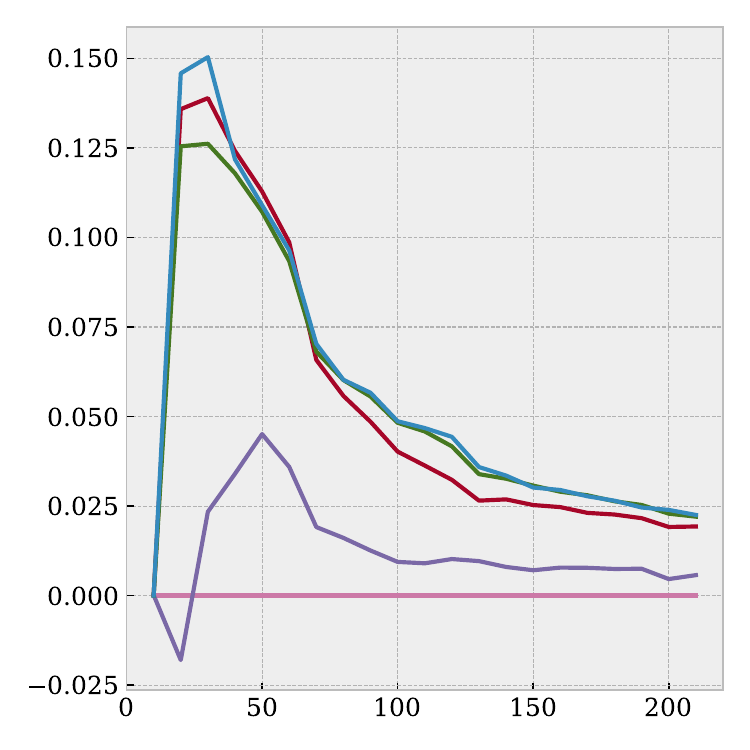}}
            \put(5, 22){\rotatebox{90}{\scriptsize Acc Difference to \Random}}
            \put(30, 7){{\scriptsize Number of Annotations}}
        \end{picture}
        \caption{CIFAR-10}
    \end{subfigure}
    \begin{subfigure}[b]{0.31\textwidth}
        \begin{picture}(100, 120)
            \put(10, 10){\includegraphics[width=\linewidth]{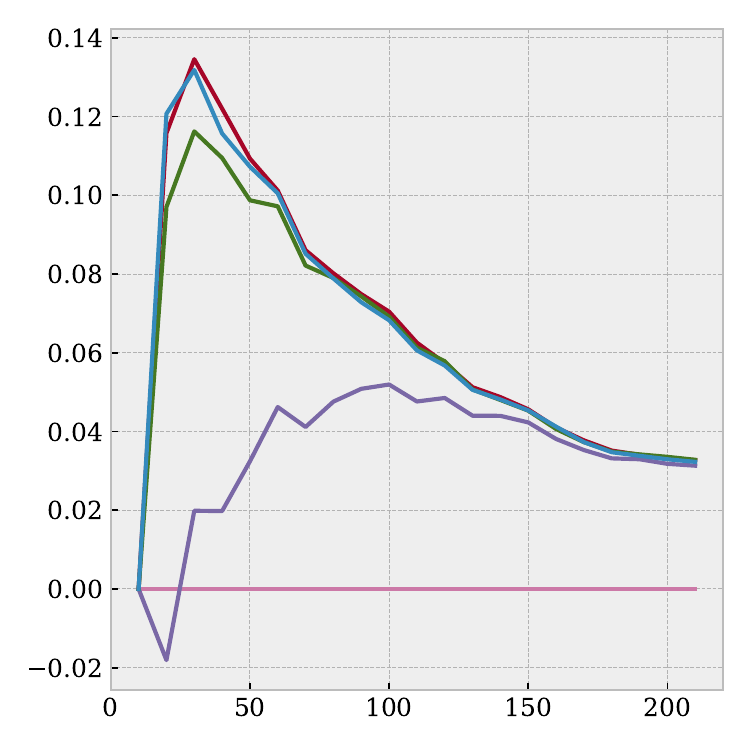}}
            \put(30, 7){{\scriptsize Number of Annotations}}
        \end{picture}
        \caption{STL-10}
    \end{subfigure}
    \begin{subfigure}[b]{0.31\textwidth}
        \begin{picture}(100, 120)
            \put(10, 10){\includegraphics[width=\linewidth]{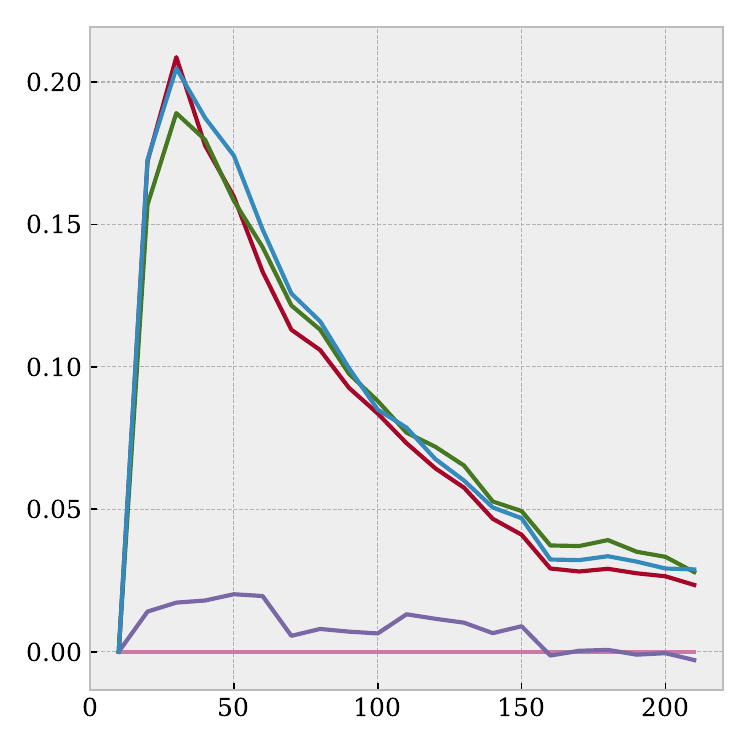}}
            \put(30, 7){{\scriptsize Number of Annotations}}
        \end{picture}
        \caption{Snacks}
    \end{subfigure}
    \caption{Accuracy improvement curves of \textsc{Bait} and its approximations.}
    \label{fig:learning_curves_ablation}
\end{figure}

\textbf{\textsc{Bait} (Binary):}
We examine the effectiveness of our second approximation discussed in Section \ref{sec:approx_gradient}, which reformulates the multi-class classification task into a binary one. Since this approximation is intended to work on a large number of classes, we also compare it to the diagonal approximation of the FIM. This approximation only calculates the main diagonal, omitting off-diagonal elements. It is frequently employed to scale the FIM to high dimensions~\cite{lecun1989optimal}, and is therefore suitable for this comparison. We refer to Appendix~\ref{app:diagonal_fim} for a more detailed explanation.
Considering Tab.~\ref{tab:ablation_likelihood}, we notice that the binary approximation yield similar (or better) accuracies compared to the original version of \textsc{Bait} across all datasets. In contrast, the diagonal FIM approximation is not able to achieve comparable performances.
Regarding the time complexity, we see that both the diagonal and binary approximation of \textsc{Bait} significantly reduce the time complexity. While the diagonal approximation is slightly slower, our approximation is able to acquire instances quicker and independent of the class number. 
This is noticeable due to (almost) constant acquisition times between Snacks and STL-10, which both have an identical number of instances but a different number of classes.
To ensure that not only the AUC values but also the behavior of the approximations are the similar, we show accuracy improvement curves in Fig.~\ref{fig:learning_curves_ablation} for \textsc{Bait} and its approximations. They show that our approximations \textsc{Bait} (Exp) and \textsc{Bait} (Binary) closely resemble the original formulation, while the diagonal approximation shows strong divergence.

\subsection{Benchmark Experiments}
Next, we present the main results on all datasets to compare the approximations of \textsc{Bait} to other state-of-the-art AL strategies. Table~\ref{tab:benchmark_results} shows the average AUC improvement over \Random of all strategies across all datasets. In the case of \textsc{Bait} (Exp), memory limitations make scaling beyond 50 classes infeasible, thus we only report results on the three datasets with less than 100 classes.
The learning curves and accuracy improvement curves for each dataset can be found in Appendix~\ref{app:learning_curves}. 

The results demonstrate that \textsc{Bait} (Binary) outperforms all state-of-the-art AL strategies on almost all datasets, except for StanfordDogs. This superiority underlines the importance of incorporating \textsc{Bait} into the evaluation of works proposing novel AL strategies. Our publicly available implementations make this incorporation easy. 
\begin{table}[t]
    \centering
    \caption{Benchmark results on image datasets. The best performing strategy is marked as bold.}
    \begin{tabular}{lclllll }
    \toprule
     & {\Random} & \Margin & \Badge & \Typiclust & \textsc{Bait} (Exp) & \textsc{Bait} (Binary)\\
    \midrule
    CIFAR-10        & {\cellcolor{lightestgray}82.05} & +2.38 & +2.55 & +3.57 & +5.43 & \bf{+5.74} \\
    STL-10          & {\cellcolor{lightestgray}85.79} & +4.42 & +3.91 & +3.70 & +6.19 & \bf{+6.45} \\
    Snacks          & {\cellcolor{lightestgray}71.93} & +3.32 & +3.08 & +7.29 & +7.50 & \bf{+8.60} \\
    CIFAR-100       & {\cellcolor{lightestgray}67.20} & +1.49 & +1.48 & +3.09 &  N/A  & \bf{+4.73} \\
    Food-101        & {\cellcolor{lightestgray}68.05} & +0.62 & +0.99 & +3.42 &  N/A  & \bf{+3.54} \\
    Flowers102      & {\cellcolor{lightestgray}56.00} & +12.05 & +10.08 & +17.64 & N/A & \bf{+18.23} \\
    StanfordDogs    & {\cellcolor{lightestgray}65.79} & +2.39 & +2.59 & \bf{+4.22} & N/A & +3.46 \\
    Tiny ImageNet   & {\cellcolor{lightestgray}63.71} & +0.67 & +1.21 & +2.52 & N/A & \bf{+3.46} \\
    Imagenet        & {\cellcolor{lightestgray}60.74} & +0.88 & +1.34 & --5.76 & N/A & \bf{+1.48} \\ 
    \bottomrule
    \end{tabular}
    \label{tab:benchmark_results}
\end{table}

\begin{figure}
    \centering
    \begin{picture}(\textwidth, 10)
    \put(50, 0){\includegraphics[width=.8\linewidth]{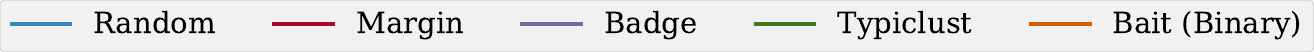}} 
    \end{picture}
    \subfloat[CIFAR-100]{
    \begin{picture}(.47\textwidth, 145)
        \put(5, 25){\rotatebox{90}{\scriptsize Accuracy Difference to \Random}}
        \put(50, 3){{\scriptsize Number of Annotations}}
        \put(10, 10){\includegraphics[width=0.47\linewidth]{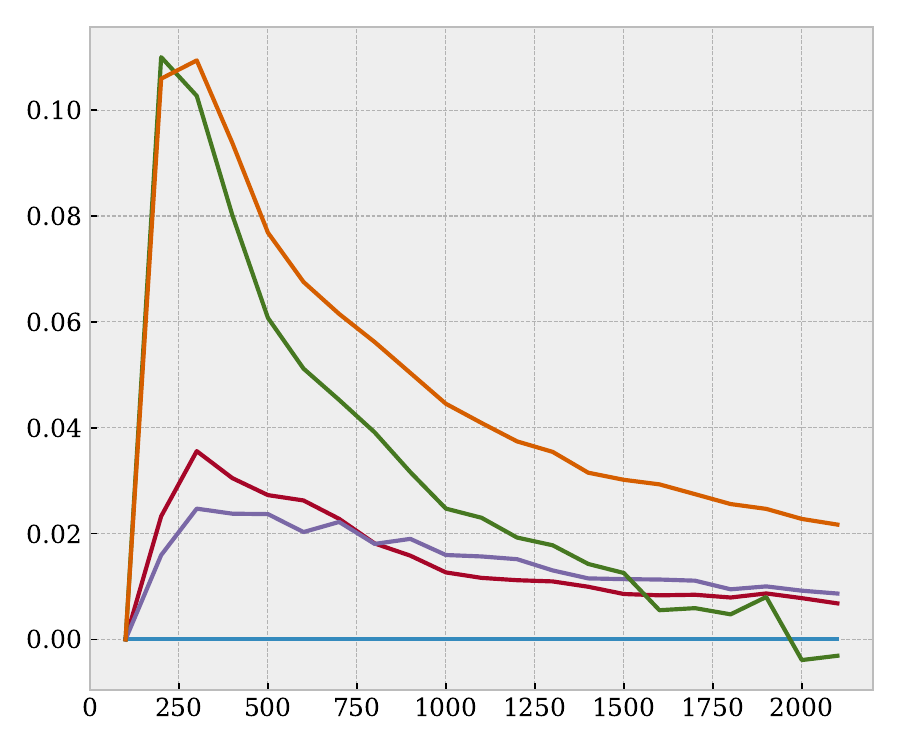}}
    \end{picture}}
    \hfill
    \subfloat[ImageNet]{
    \begin{picture}(0.47\textwidth, 145)
        \put(5, 25){\rotatebox{90}{\scriptsize Accuracy Difference to \Random}}
        \put(50, 3){{\scriptsize Number of Annotations}}
        \put(10, 10){\includegraphics[width=0.47\linewidth]{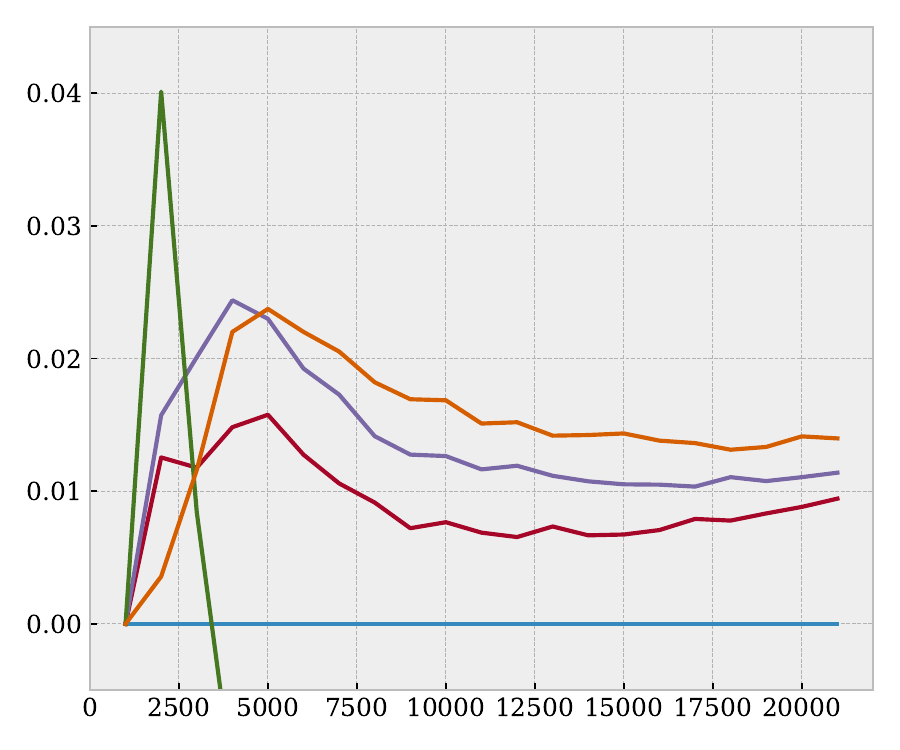}}    
    \end{picture}
    }
    \caption{Accuracy improvement curves of state-of-the-art strategies.}
    \label{fig:benchmark_learning_curves}
\end{figure}

In Fig.~\ref{fig:benchmark_learning_curves}, we exemplary show accuracy improvement curves for CIFAR-100 and ImageNet. For CIFAR-100, \textsc{Bait} and \Typiclust exhibit superior performance in the early stages of AL. This is also corroborated when examining the learning curves of all other datasets. Most likely, this is due to a more effective diverse selection, which have shown to be more beneficial in early AL cycles~\cite{hacohen2022active}. As the process progresses, we see that \Badge and \Margin catch up. At the cycle, the accuracy of \Typiclust becomes worse than the accuracy of \Badge, \Margin, and even \Random. This demonstrates that \Typiclust's extensive focus on a diverse selection can harm the accuracy in AL. As studied by~\cite{hacohen2023how}, there exists a transition where a stronger focus on difficult instances is more beneficial than forcing diversity. Considering our approximation, we see that \textsc{Bait} performs superior to the other strategies, suggesting that its selection is efficient during the entire AL process. 
For ImageNet, we see that \textsc{Bait} provides the strongest improvement compared to the other strategies. Especially, the results demonstrate that \Typiclust is not suitable for a scenario with a large number of classes and acquisition sizes.
By examining the other learning curves, we found that \textsc{Bait} occasionally leads to a worse final accuracy than \Badge. We suppose that a more accurate approximation of the FIM becomes more important in later AL cycles, and therefore \textsc{Bait} (Binary) does not exploit the full potential in approximating \textsc{Bait}. We leave further investigation of this for future work.

\section{Conclusion}\label{sec:conclusion}
In this article, we explored the time and space complexity of \textsc{Bait} and addressed its scalability issues when applied to tasks with a large number of classes. We presented two methods to reduce its time and space complexity by approximating the FIM.
Our first method, \textsc{Bait} (Exp), modifies \textsc{Bait}'s original  formulation by taking the expectation over the most probable classes, reducing the time complexity from $\mathcal{O}(K^3D^2)$ to $\mathcal{O}(c K^2 D^2)$.
Our second method, \textsc{Bait} (Binary), considers a binary classification task, leading to an alternative likelihood for gradient computations, considerably reducing the time complexity to~$\mathcal{O}(D^2)$.
This adaptation enables \textsc{Bait}'s usage on large-scale datasets, such as ImageNet.
An extensive evaluation across nine image datasets demonstrates that our approximations perform similarly to the original formulation of \textsc{Bait}, and outperform existing state-of-the-art AL strategies in terms of accuracy.

For practitioners working with image datasets, we suggest using the \Bait (Binary) method. Our research mainly focused on image data and demonstrated that the approximation is effective in this context. For other data modalities, such as text or tabular data, we recommend using \Bait (Exp), as it uses a biased estimator of the FIM and is therefore closer to \Bait's original design. Further, we suggest setting $c = 2$ since focusing on the two most probable classes yields well working approximations. Our implementation is publicly available and provides an easy way to integrate \Bait into existing frameworks.

For future work, we plan to further validate the effectiveness of our approximations by evaluating them on different data modalities. For text data in particular, we are interested in the effectiveness of combining \Bait with models like Bert~\cite{devlin2019bert}. For image data, we plan to evaluate our methods even more extensively. We want to ensure that our approximations behave similarly to the original version of \Bait. For this purpose, we will employ more datasets and perform statistical tests for verification.

\bibliographystyle{splncs04}
\bibliography{ref}

\newpage
\appendix
\section{Experimental Setup}\label{app:graphical_abstract}
The graphical abstract is based on the same experimental setup as the main experiments. For the Active Learning (AL) setting, we use the CIFAR-10 dataset consisting of ten classes. We start with 0 labeled instances and select 10 instances per cycle for a total of 21 cycles. In the first cycle, all instances are sampled randomly. We employ the DINOv2-ViT-S/14 model~\cite{oquab2023dinov2} with a fully connected linear classification layer. In each AL cycle, we randomly initialize the classification head for 200 epochs and employ the Rectified Adam optimizer~\cite{liu2019variance}. We use a training batch size of 128, a learning rate of 0.2, and weight decay of 0.0001. In addition, we utilize a cosine annealing learning rate scheduler.

\section{Dataset Description}\label{app:datasets}
Here, we describe the datasets used in the main experiments in more detail.
\noindent
\textbf{CIFAR}~\cite{krizhevsky2009learning} consists of 60,000 colored images with a low $32 \times 32$ resolution. There is a predetermined split of 50,000 images as training samples and 10,000 images as test samples. Two variants of this dataset exist, distinguished by their differing number of classes. \textbf{CIFAR-10} is a coarse-grained task with broad classes such as automobile, dog, airplane, and ship. The classes are mutually exclusive meaning there is no overlap between, e.g., automobiles and trucks. \textbf{CIFAR-100}, containing 100 classes, provides more fine-grained differentiation, such as bicycle, apple, television, and roses.
\noindent\textbf{Food-101}~\cite{bossard2014food} consists of 101 food categories with 750 training and 250 test images per category, making a total of 101,000 images. \textbf{Snacks} is a smaller dataset, containing 20 different classes of snack foods, and is split into 5,000 training images and 1,000 test images. \textbf{Flowers-102}~\cite{nilsback2008automated} consists of 102 flower categories, commonly occurring in the United Kingdom. Each class consists of 40 to 258 images, making it an imbalanced dataset with a total of 8,189 images. The images have large scale, pose, and light variations. In addition, some categories have large variations within the category and several very similar categories. \textbf{StanfordDogs}~\cite{zhao2020universal} contains 20,580 images of 120 classes of dogs from around the world. It is partitioned into 12,000 images for training and 8,580 images for testing.
\textbf{ImageNet}~\cite{deng2009imagenet} is a large-scale image database with 14,197,122 labeled images. They are organized by the semantic hierarchy of WordNet. Images vary in size and are, therefore, usually cropped to 224×224. A common subset is the ImageNet1k dataset, containing 1,281,167 training images, 100,000 test images, and an additional 50,000 validation images, which can be classified into 1,000 classes. In our work, we refer to ImageNet1k as ImageNet. \textbf{Tiny ImageNet}~\cite{le2015tiny} is a scaled-down subset of ImageNet. It consists of 200 classes, and for each class, it provides 500 training images, 50 validation images, and 50 test images. All images have been downsampled to 64×64. \textbf{STL-10}~\cite{wang2016unsupervised} is another derivation of ImageNet commonly used to evaluate algorithms of unsupervised feature learning. It contains 100,000 unlabeled images and 13,000 labeled images. The labeled images are divided into 5,000 training images and 8,000 test images. All images have been downsampled to 96×96.

\section{Diagonal Approximation of the Fisher Information Matrix}\label{app:diagonal_fim}
A common approximation of the Fisher information matrix (FIM) is to only consider the diagonal elements~\cite{lecun1989optimal}. This simplifies the outer product considerably reducing the time and space complexity: 
\begin{align}
    \boldsymbol{I}(\x;\params) \approx \mathbb{E}_\y \left[\diag(\nabla_{\params} \ln p(\y | \x, \params) \odot \nabla_{\params} \ln p(\y | \x, \params))\right]
\end{align}
where $\odot$ is the Hadamard product.
This basically omits the off-diagonal elements, which represent the correlations between parameters. This can lead to a less accurate understanding of the parameter space, especially in models where parameter interactions are significant.

\begin{table}[!t]
\setlength{\tabcolsep}{1pt}
\centering
\caption{Overview of AL-related hyperparameters for each dataset.}
\begin{tabular}{|c|c|c|c|}
\toprule
\textbf{Dataset}  & \textbf{Initial Pool Size} & \textbf{Acquisition Size} & \textbf{Number of Acquisitions} \\ 
\midrule
CIFAR10         & 10   & 10   &  20 \\
CIFAR100        & 100  & 100  &  20 \\
Food-101        & 100  & 100  &  20 \\
Flowers-102     & 50   & 50   &  20 \\
StanfordDogs    & 100  & 100  &  20 \\
Tiny ImageNet   & 200  & 200  &  20 \\
ImageNet        & 1000 & 1000 &  20 \\
\bottomrule
\end{tabular}
\label{tab:al_setting}
\end{table}
\section{Active Learning Settings}\label{app:al_settings}
In Tab.~\ref{tab:al_setting}, we list AL related settings. We derived these values by qualitatively investigating the learning curves of \textsc{Random} sampling in initial experiments with a 10\% split of the training dataset. This way, we try to ensure convergence of learning curves across strategies, which is necessary for the area under the learning curve of a strategy to be representative.

\section{Learning Curves}\label{app:learning_curves}
All learning curves from the main experiments are depicted in Fig.~\ref{fig:learning_curves_1},~\ref{fig:learning_curves_2}, and~\ref{fig:learning_curves_3}. For more details, we refer to our implementation.
\newcommand\fsize{0.45\textwidth}
\begin{figure}
    \centering
    
    \begin{picture}(300, 20)
    \put(30, 0){\includegraphics[width=0.7\textwidth]{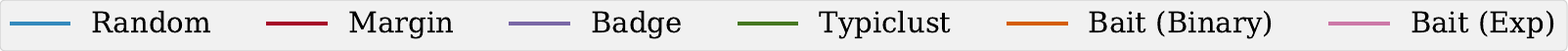}}
    \end{picture}
    \vspace{1em}
    
    \subfloat[Learning Curve]{
        \begin{picture}(150, 140)
        \put(-5, 10){\includegraphics[width=\fsize]{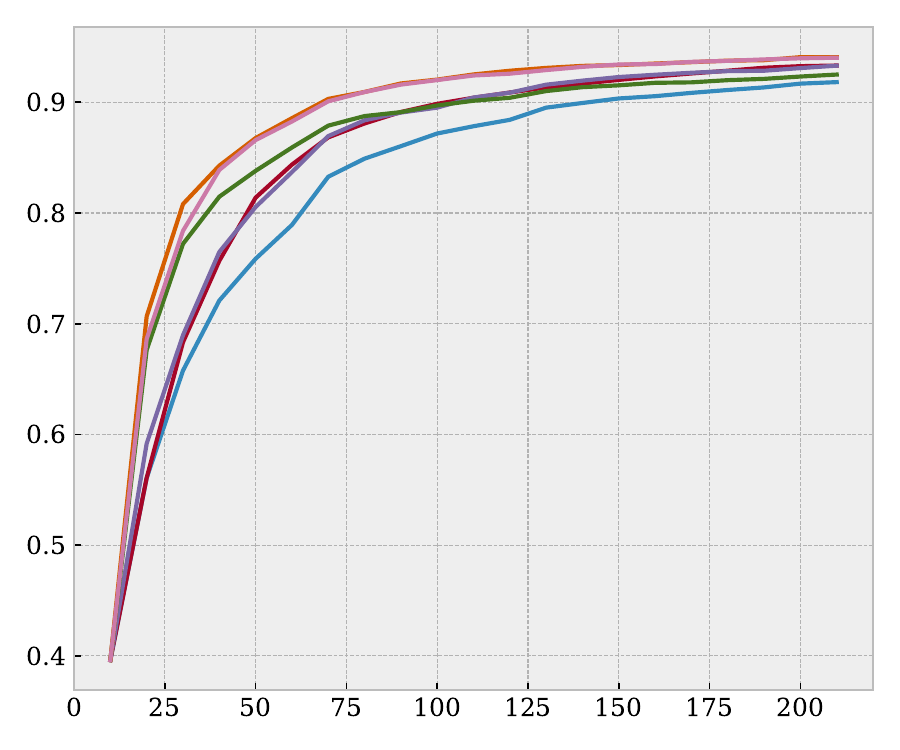}}
        \put(-10, 60){\rotatebox{90}{\scriptsize Accuracy}}
        \put(40, 7){{\scriptsize Number of Annotations}}
        \end{picture}
    }
    \quad
    \subfloat[Accuracy Difference to \textsc{Random}]{
        \begin{picture}(150, 140)
        \put(-5, 10){\includegraphics[width=\fsize]{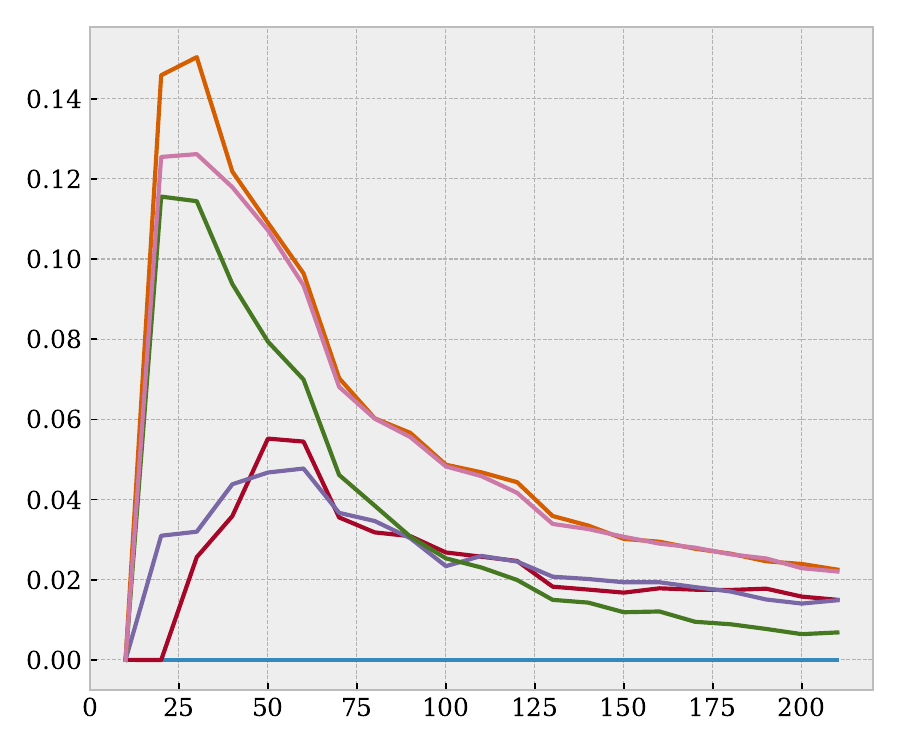}}
        \put(-10, 23){\rotatebox{90}{\scriptsize Accuracy Difference to Random}}
        \put(40, 7){{\scriptsize Number of Annotations}}
        \put(-30, 138){{\bf CIFAR-10}}
        \end{picture}
    } 
    \vspace{1em}
    
    \subfloat[Learning Curve]{
        \begin{picture}(150, 140)
        \put(-5, 10){\includegraphics[width=\fsize]{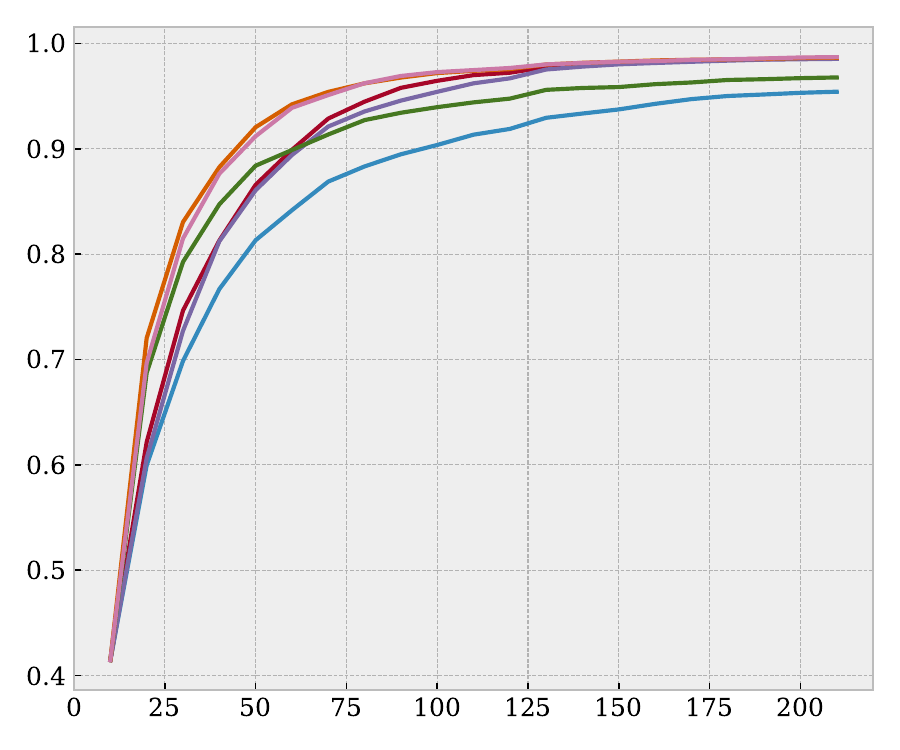}}
        \put(-10, 60){\rotatebox{90}{\scriptsize Accuracy}}
        \put(40, 7){{\scriptsize Number of Annotations}}
        \end{picture}
    }
    \quad
    \subfloat[Accuracy Difference to \textsc{Random}]{
        \begin{picture}(150, 140)
        \put(-5, 10){\includegraphics[width=\fsize]{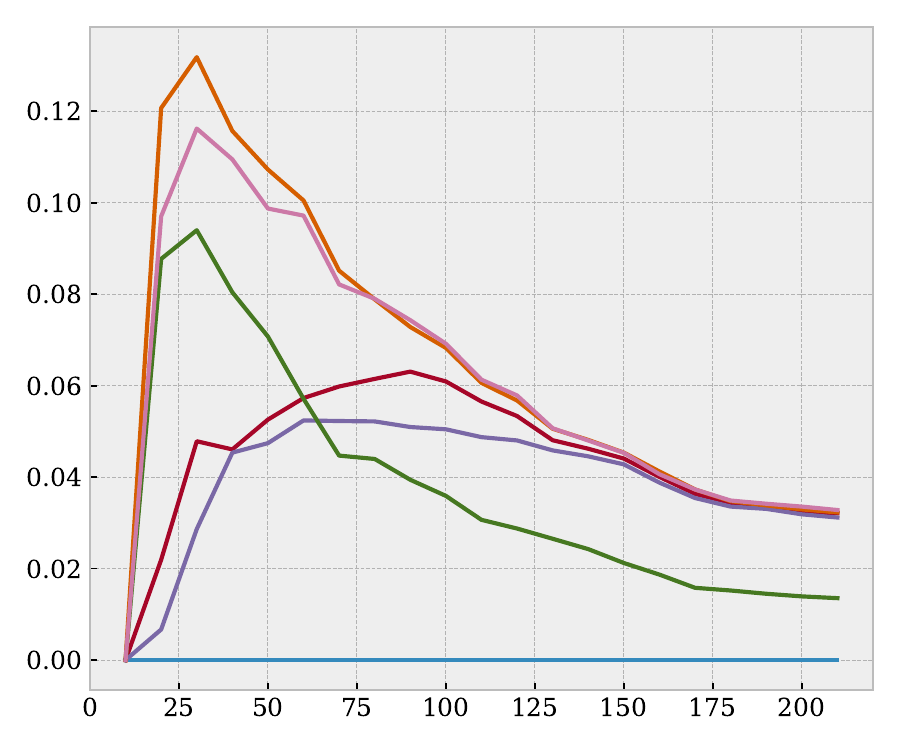}}
        \put(-10, 23){\rotatebox{90}{\scriptsize Accuracy Difference to Random}}
        \put(40, 7){{\scriptsize Number of Annotations}}
        \put(-23, 138){{\bf STL-10}}
        \end{picture}
    } 
    \vspace{1em}
    
    \subfloat[Learning Curve]{
        \begin{picture}(150, 140)
        \put(-5, 10){\includegraphics[width=\fsize]{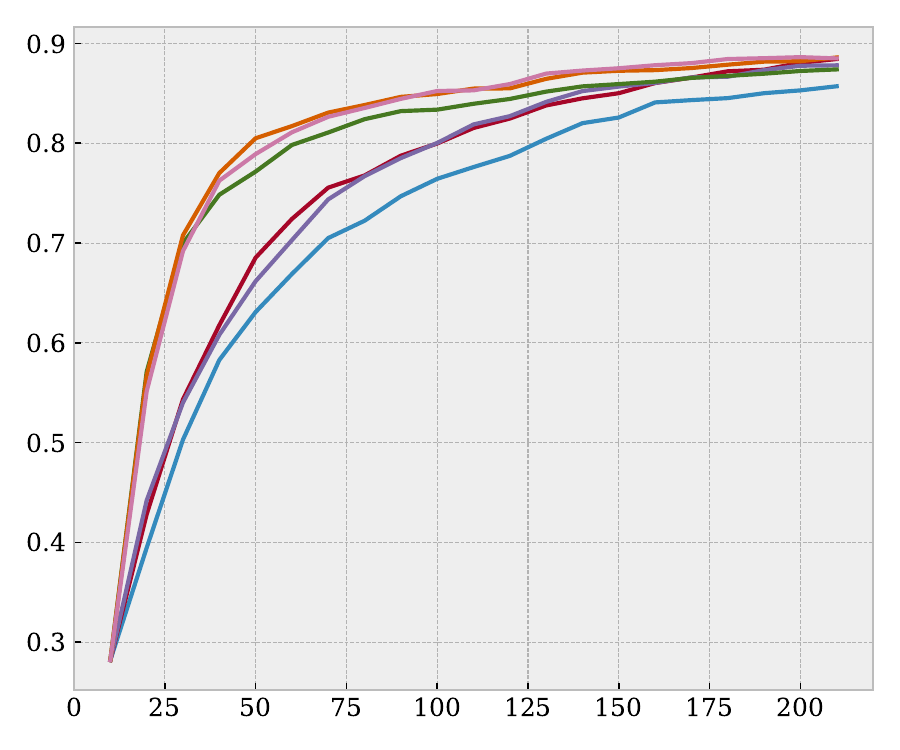}}
        \put(-10, 60){\rotatebox{90}{\scriptsize Accuracy}}
        \put(40, 7){{\scriptsize Number of Annotations}}
        \end{picture}
    }
    \quad
    \subfloat[Accuracy Difference to \textsc{Random}]{
        \begin{picture}(150, 140)
        \put(-5, 10){\includegraphics[width=\fsize]{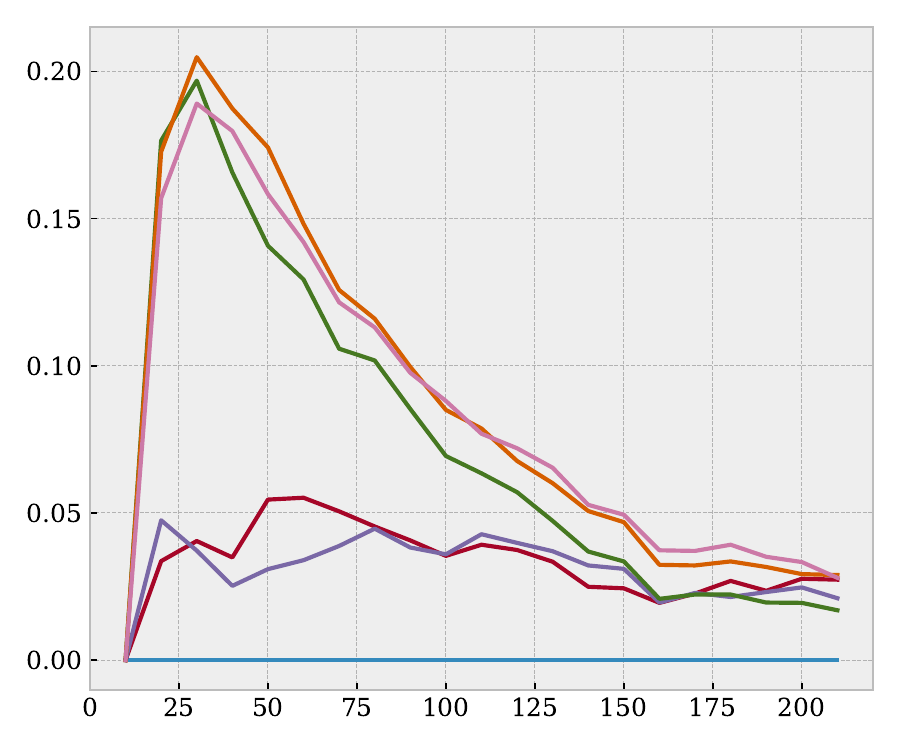}}
        \put(-10, 23){\rotatebox{90}{\scriptsize Accuracy Difference to Random}}
        \put(40, 7){{\scriptsize Number of Annotations}}
        \put(-22, 138){{\bf Snacks}}
        \end{picture}
    } 
   
    \caption{Accuracy improvement curves of state-of-the-art strategies.}
    \label{fig:learning_curves_1}
\end{figure}

\begin{figure}
    \centering
    \begin{picture}(300, 20)
    \put(30, 0){\includegraphics[width=0.7\textwidth]{figures/benchmark/cifar100_benchmark_legend.pdf}}
    \end{picture}
    \vspace{1em}
    
    \subfloat[Learning Curve]{
        \begin{picture}(150, 140)
        \put(-5, 10){\includegraphics[width=\fsize]{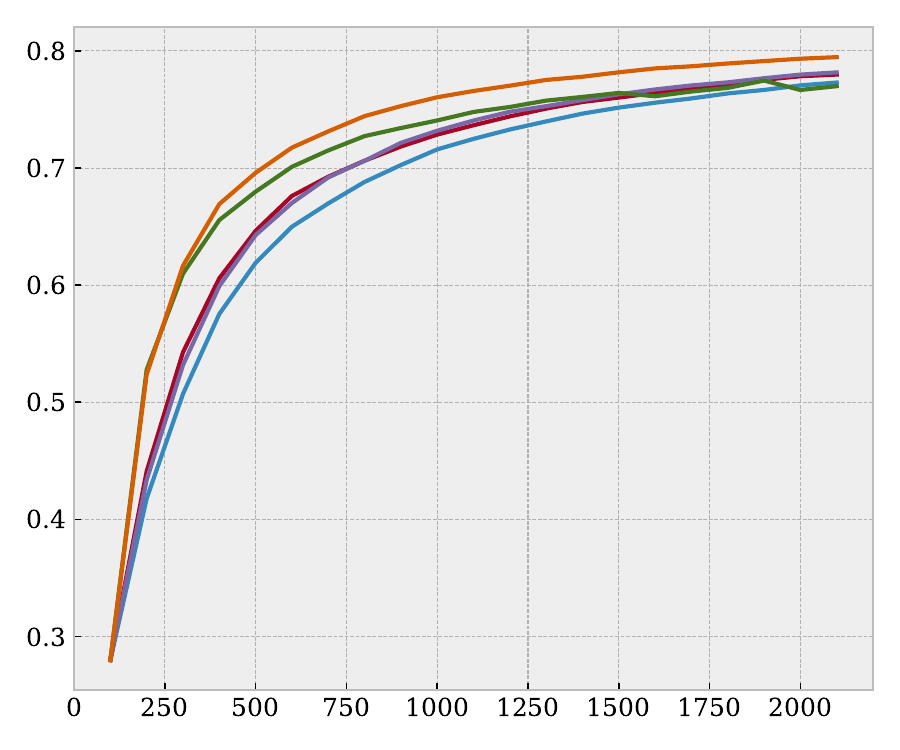}}
        \put(-10, 60){\rotatebox{90}{\scriptsize Accuracy}}
        \put(40, 7){{\scriptsize Number of Annotations}}
        \end{picture}
    }
    \quad
    \subfloat[Accuracy Difference to \textsc{Random}]{
        \begin{picture}(150, 140)
        \put(-5, 10){\includegraphics[width=\fsize]{figures/benchmark/cifar100_benchmark_lc_diff.pdf}}
        \put(-10, 23){\rotatebox{90}{\scriptsize Accuracy Difference to Random}}
        \put(40, 7){{\scriptsize Number of Annotations}}
        \put(-30, 138){{\bf CIFAR-100}}
        \end{picture}
    } 
    \vspace{1em}
    
    \subfloat[Learning Curve]{
        \begin{picture}(150, 140)
        \put(-5, 10){\includegraphics[width=\fsize]{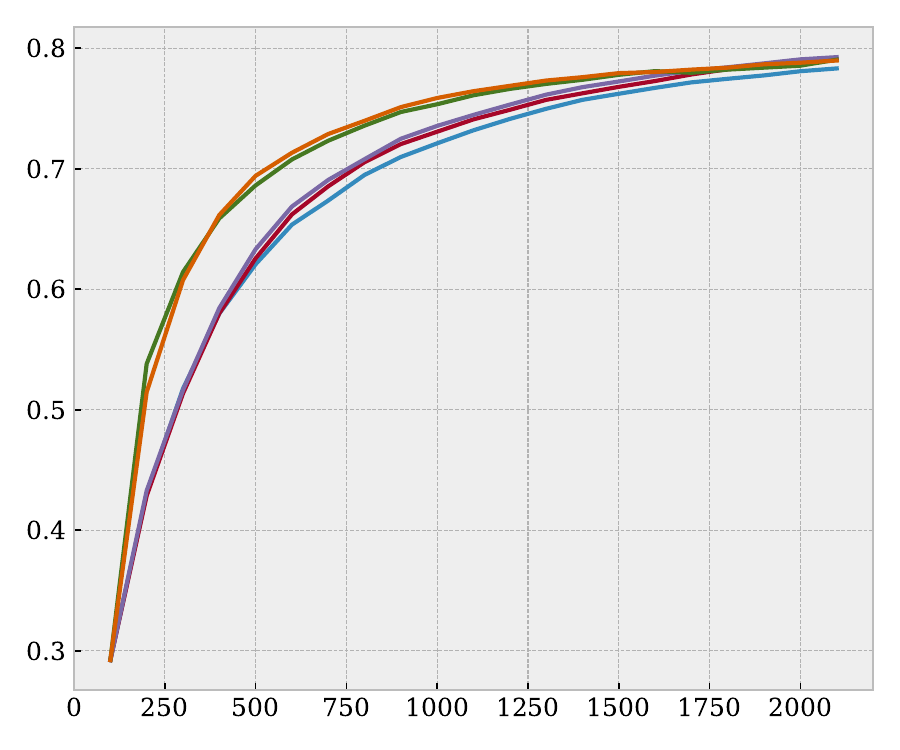}}
        \put(-10, 60){\rotatebox{90}{\scriptsize Accuracy}}
        \put(40, 7){{\scriptsize Number of Annotations}}
        \end{picture}
    }
    \quad
    \subfloat[Accuracy Difference to \textsc{Random}]{
        \begin{picture}(150, 140)
        \put(-5, 10){\includegraphics[width=\fsize]{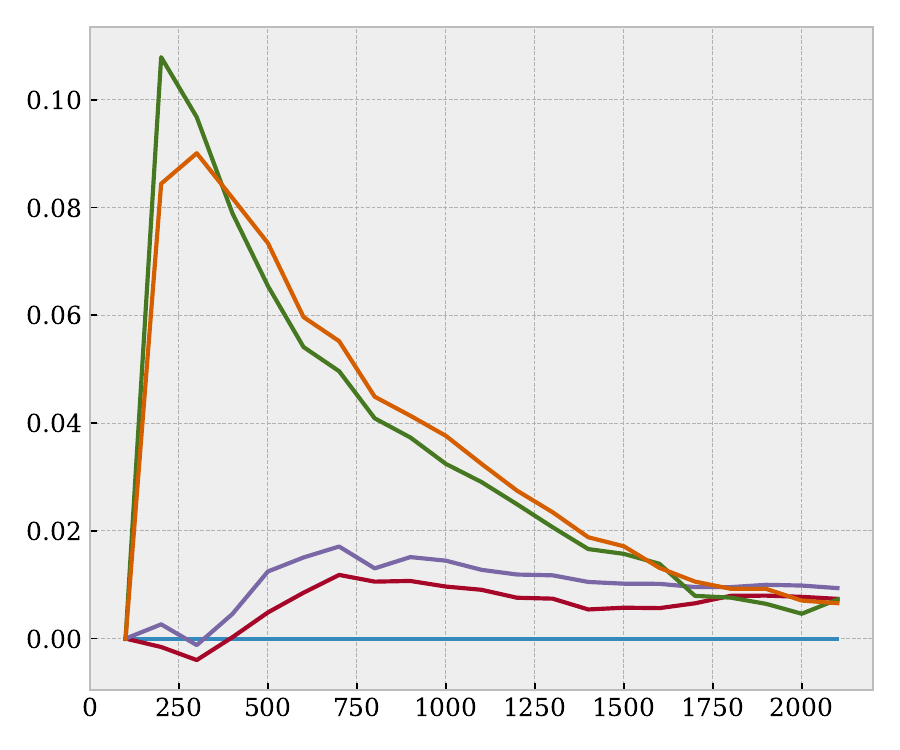}}
        \put(-10, 23){\rotatebox{90}{\scriptsize Accuracy Difference to Random}}
        \put(40, 7){{\scriptsize Number of Annotations}}
        \put(-25, 138){{\bf Food-101}}
        \end{picture}
    } 
    \vspace{1em}
    
    \subfloat[Learning Curve]{
        \begin{picture}(150, 140)
        \put(-5, 10){\includegraphics[width=\fsize]{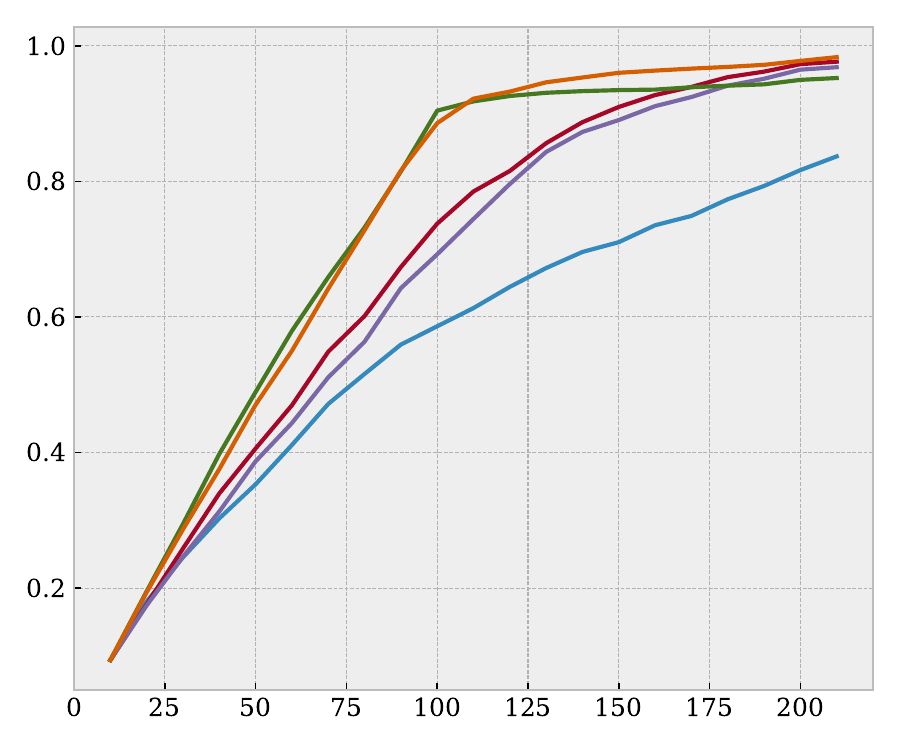}}
        \put(-10, 60){\rotatebox{90}{\scriptsize Accuracy}}
        \put(40, 7){{\scriptsize Number of Annotations}}
        \end{picture}
    }
    \quad
    \subfloat[Accuracy Difference to \textsc{Random}]{
        \begin{picture}(150, 140)
        \put(-5, 10){\includegraphics[width=\fsize]{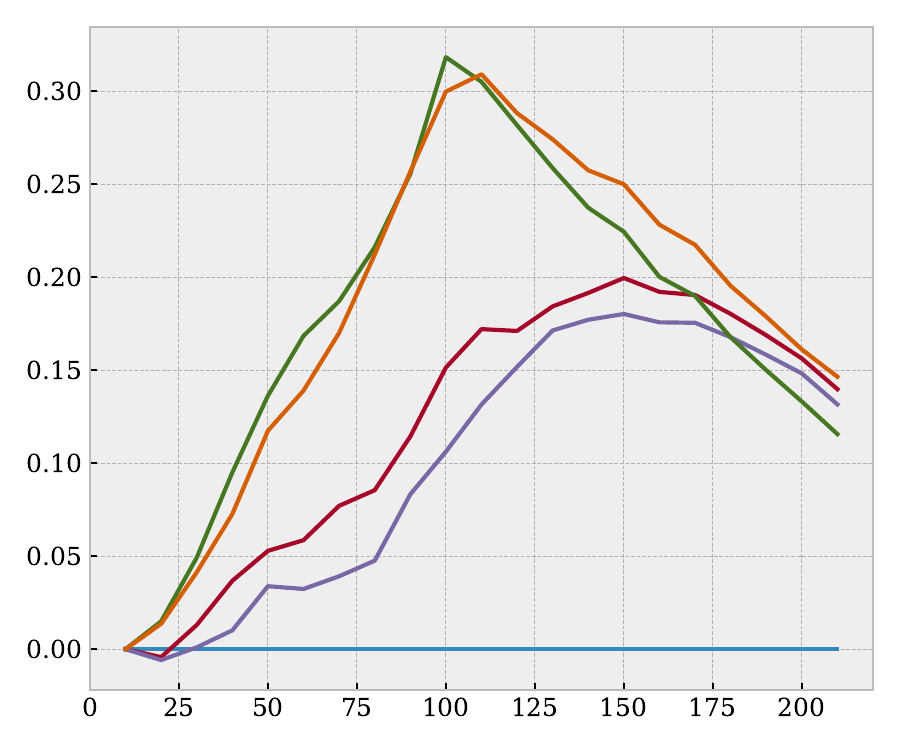}}
        \put(-10, 23){\rotatebox{90}{\scriptsize Accuracy Difference to Random}}
        \put(40, 7){{\scriptsize Number of Annotations}}
        \put(-30, 138){{\bf Flowers-102}}
        \end{picture}
    } 
    \caption{Accuracy improvement curves of state-of-the-art strategies.}
    \label{fig:learning_curves_2}
\end{figure}

\begin{figure}
    \centering
    \begin{picture}(300, 20)
    \put(50, 0){\includegraphics[width=0.7\textwidth]{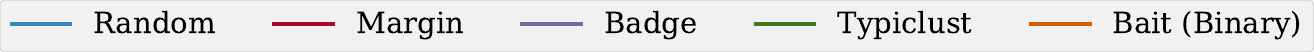}}
    \end{picture}
    \vspace{1em}
    \\
    \subfloat[Learning Curve]{
        \begin{picture}(150, 140)
        \put(-5, 10){\includegraphics[width=\fsize]{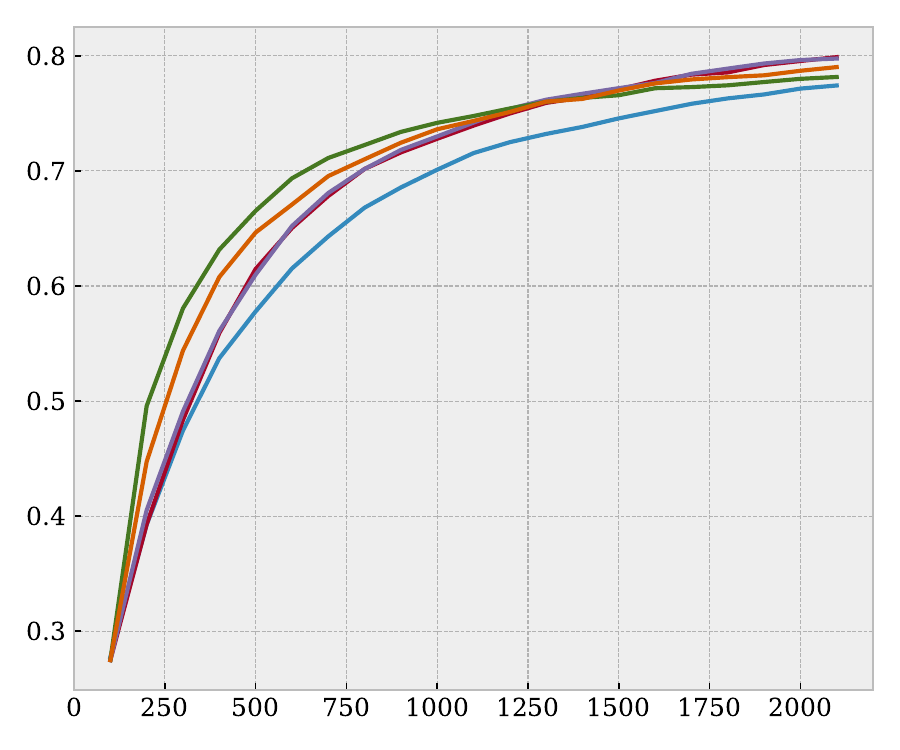}}
        \put(-10, 60){\rotatebox{90}{\scriptsize Accuracy}}
        \put(40, 7){{\scriptsize Number of Annotations}}
        \end{picture}
    }
    \quad
    \subfloat[Accuracy Difference to \textsc{Random}]{
        \begin{picture}(150, 140)
        \put(-5, 10){\includegraphics[width=\fsize]{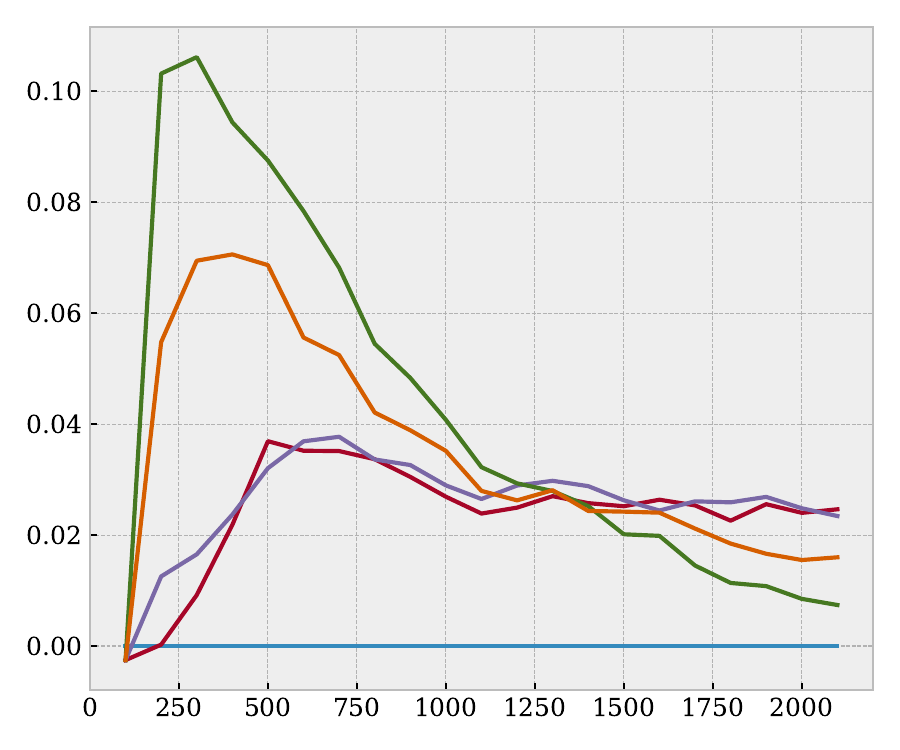}}
        \put(-10, 23){\rotatebox{90}{\scriptsize Accuracy Difference to Random}}
        \put(40, 7){{\scriptsize Number of Annotations}}
        \put(-37, 138){{\bf StandfordDogs}}
        \end{picture}
    } 
    \vspace{1em}
    
    \subfloat[Learning Curve]{
        \begin{picture}(150, 140)
        \put(-5, 10){\includegraphics[width=\fsize]{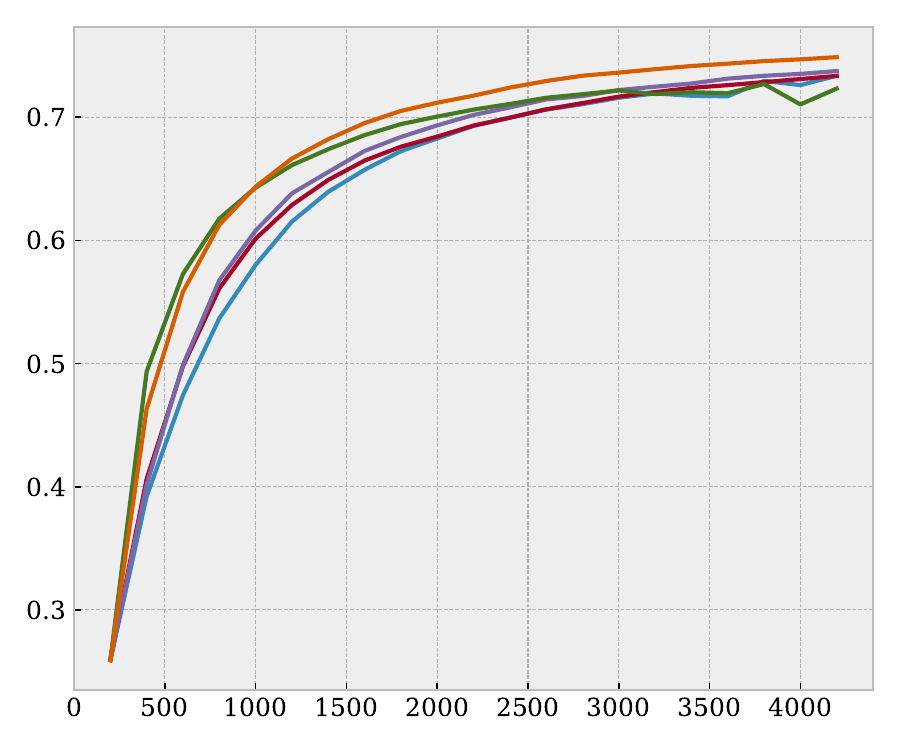}}
        \put(-10, 60){\rotatebox{90}{\scriptsize Accuracy}}
        \put(40, 7){{\scriptsize Number of Annotations}}
        \end{picture}
    }
    \quad
    \subfloat[Accuracy Difference to \textsc{Random}]{
        \begin{picture}(150, 140)
        \put(-5, 10){\includegraphics[width=\fsize]{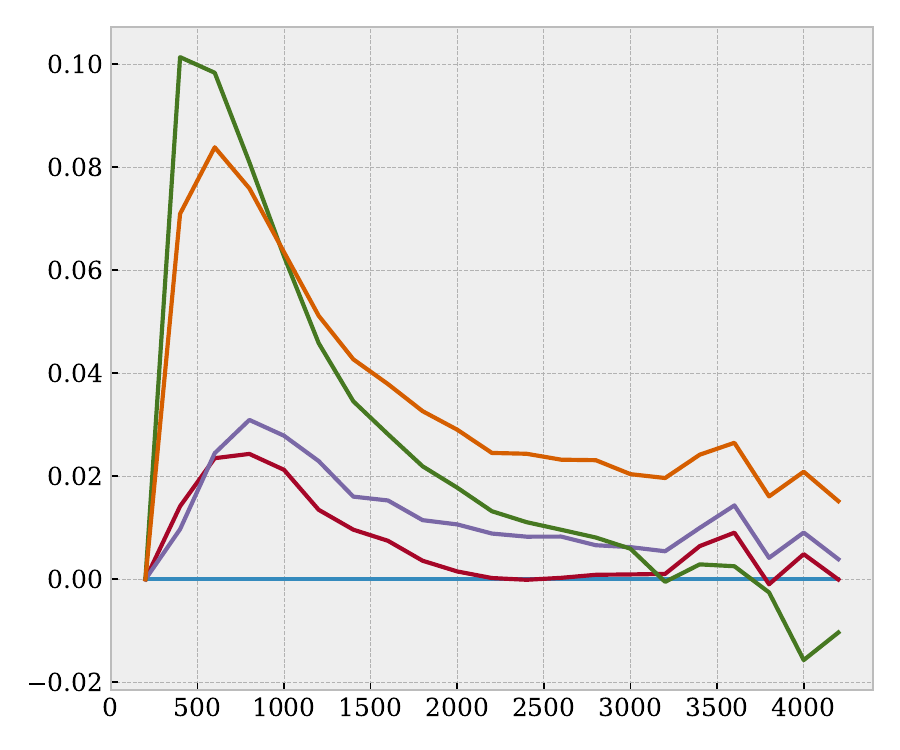}}
        \put(-10, 23){\rotatebox{90}{\scriptsize Accuracy Difference to Random}}
        \put(40, 7){{\scriptsize Number of Annotations}}
        \put(-32, 138){{\bf TinyImagenet}}
        \end{picture}
    } 
    \vspace{1em}
    
    \subfloat[Learning Curve]{
        \begin{picture}(150, 140)
        \put(-5, 10){\includegraphics[width=\fsize]{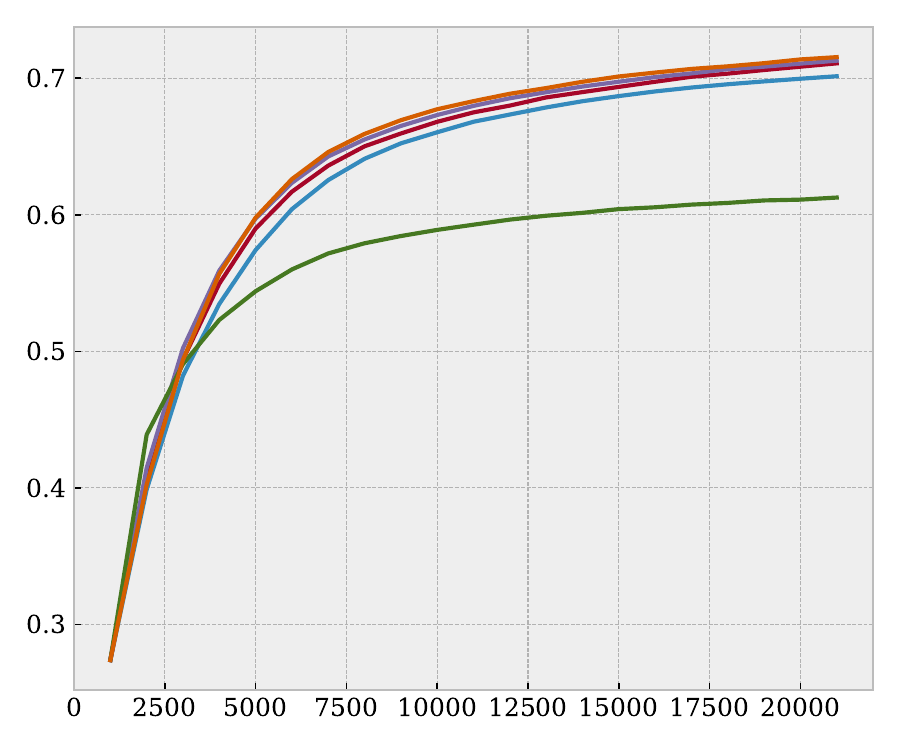}}
        \put(-10, 60){\rotatebox{90}{\scriptsize Accuracy}}
        \put(40, 7){{\scriptsize Number of Annotations}}
        \end{picture}
    }
    \quad
    \subfloat[Accuracy Difference to \textsc{Random}]{
        \begin{picture}(150, 140)
        \put(-5, 10){\includegraphics[width=\fsize]{figures/benchmark/imagenet_benchmark_lc_diff.pdf}}
        \put(-10, 23){\rotatebox{90}{\scriptsize Accuracy Difference to Random}}
        \put(40, 7){{\scriptsize Number of Annotations}}
        \put(-25, 138){{\bf Imagenet}}
        \end{picture}
    } 
    \caption{Accuracy improvement curves of state-of-the-art strategies.}
    \label{fig:learning_curves_3}
\end{figure}

\end{document}